%% file: main.tex
\documentclass[10pt,twocolumn,letterpaper]{article}

\usepackage[pagenumbers]{cvpr} 
\usepackage[utf8]{inputenc} 
\usepackage[T1]{fontenc}    

\usepackage{url}            
\usepackage{booktabs}       
\usepackage{amsfonts}       
\usepackage{nicefrac}       
\usepackage{microtype}      
\usepackage{lipsum}
\usepackage{natbib}
\usepackage{graphicx}
\usepackage{amsmath}
\usepackage[table]{xcolor}
\usepackage{tabularx}
\usepackage{siunitx}
\usepackage{colortbl}
\usepackage{makecell}
\usepackage{threeparttable}
\usepackage{multirow}
\usepackage{adjustbox} 
\usepackage{array}
\usepackage{graphicx}
\usepackage{caption}
\usepackage{subcaption}
\usepackage{comment}
\usepackage{fontawesome5}  

\usepackage{xcolor}
\usepackage{fontawesome5}
\definecolor{ModelGreen}{RGB}{213,232,212}
\newcolumntype{L}[1]{>{\raggedright\arraybackslash}p{#1}}

\input{preamble}
\definecolor{cvprblue}{rgb}{0.21,0.49,0.74}
\definecolor{LightGray}{gray}{0.9}       
\definecolor{PaleBlue}{rgb}{0.9, 0.95, 1.0} 
\definecolor{MintGreen}{rgb}{0.9, 1.0, 0.9}  
\usepackage[pagebackref,breaklinks,colorlinks,allcolors=cvprblue]{hyperref}
\usepackage{hyperref}
\usepackage{fontawesome5}
\def\paperID{****} 
\def\confName{CVPR}
\def\confYear{2026}

\title{MedMO: Grounding and Understanding Multimodal Large Language Model for Medical Images}

\author{
Ankan Deria*,
Komal Kumar*,
Adinath Madhavrao Dukre,
Eran Segal,
Salman Khan,
Imran Razzak \\
Mohamed bin Zayed University of Artificial Intelligence \\
{\tt\small \{ankan.deria, komal.kumar\}@mbzuai.ac.ae} \\[4pt]
\small
\raisebox{-0.15em}{\includegraphics[height=0.9em]{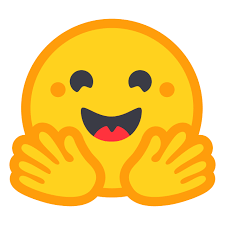}}~%
\textbf{Models:}~\href{https://huggingface.co/collections/MBZUAI/medmo}{\texttt{\textcolor{teal}{huggingface.co/collections/MBZUAI/medmo}}}~~ \\[2pt]
\small
{\fontsize{10}{10}\selectfont\faGithub}~\textbf{GitHub:}~\href{https://github.com/genmilab/MedMO}{\texttt{\textcolor{teal}{github.com/genmilab/MedMO}}}~~ \\[2pt]
\small
{\fontsize{10}{10}\selectfont\faGlobe}~\textbf{Project Page:}~\href{https://genmilab.github.io/MedMO-Page/}{\texttt{\textcolor{teal}{genmilab.github.io/MedMO-Page}}}
}

\begin{document}

\twocolumn[{%
\renewcommand\twocolumn[1][]{#1}%
\maketitle
\begin{center}
    \centering
    \includegraphics[width=1.0\textwidth]{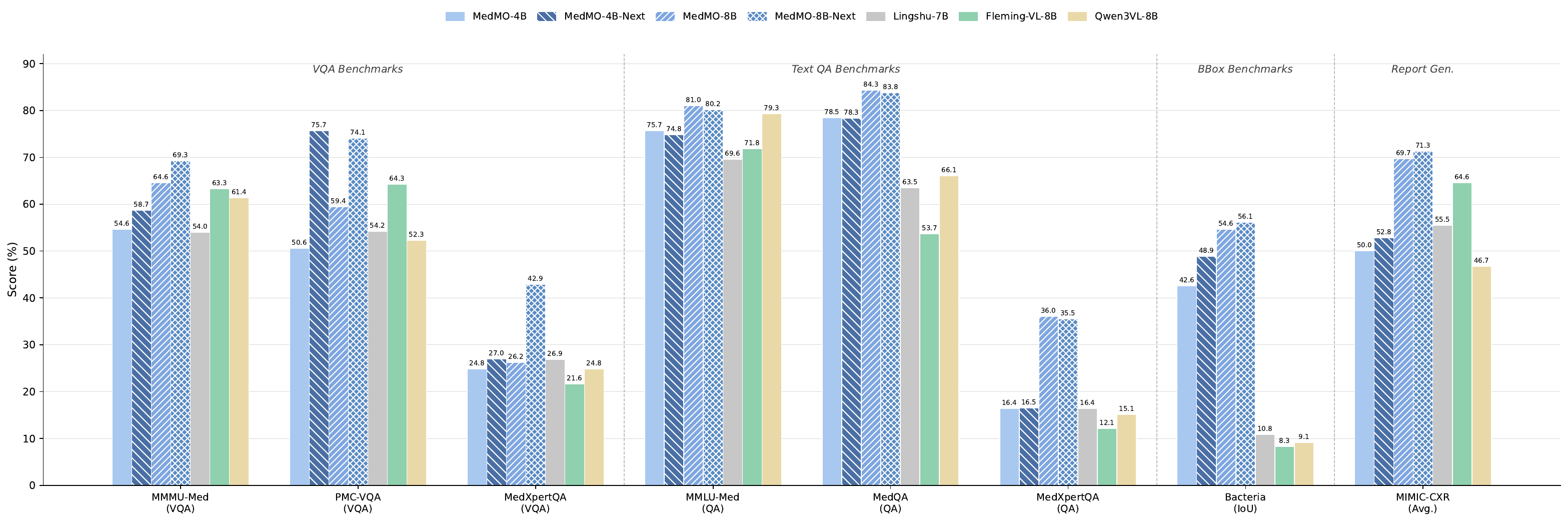}
    \vspace{-10pt}
    \captionof{figure}{\small
Benchmark performance of \textbf{MedMO-4B} and \textbf{MedMO-8B} variants (base and \textit{Next}) across medical VQA, QA, grounding, and report generation. 
\textbf{MedMO-8B-Next} consistently leads all comparisons, outperforming Fleming-VL-8B by \textcolor{ForestGreen}{+6.0\%} on MMMU-Med (\textcolor{ForestGreen}{69.3\%} vs. 63.3\%), \textcolor{ForestGreen}{+9.8\%} on PMC-VQA (\textcolor{ForestGreen}{74.1\%} vs. 64.3\%), \textcolor{ForestGreen}{+8.4\%} on MMLU-Med (\textcolor{ForestGreen}{80.2\%} vs. 71.8\%), \textcolor{ForestGreen}{+17.7\%} on MedQA (\textcolor{ForestGreen}{83.8\%} vs. 66.1\%), \textcolor{ForestGreen}{+15.8\%} on MIMIC-CXR (\textcolor{ForestGreen}{71.3\%} vs. 55.5\%), and \textcolor{ForestGreen}{+47.0} IoU on Bacteria grounding (\textcolor{ForestGreen}{56.1} vs. 9.1).
\textbf{MedMO-4B-Next} remains competitive despite its smaller scale, surpassing Fleming-VL-8B on most benchmarks.
Overall, MedMO-8B-Next achieves the best average scores (VQA: \textcolor{ForestGreen}{72.7\%}, QA: \textcolor{ForestGreen}{60.1\%}) against Fleming-VL-8B (VQA: \textcolor{BrickRed}{66.1\%}, QA: \textcolor{BrickRed}{45.7\%}), while even the compact MedMO-4B-Next (VQA: \textcolor{ForestGreen}{68.5\%}, QA: \textcolor{ForestGreen}{55.0\%}) outperforms Fleming across both categories, and MedMO-8B (VQA: 63.2\%, QA: \textcolor{ForestGreen}{61.3\%}) demonstrates strong QA reasoning despite a lower VQA average.
}
    \label{fig:intro_fig}
\end{center}%
}]
\input{sec/0_abstract}
\input{sec/1_intro}

\input{sec/2_methodology}

\input{sec/3_experiments}

{
    \small
    \bibliographystyle{ieeenat_fullname}
    \bibliography{main}
}

\input{sec/X_suppl}


\end{document}

%% file: sec/0_abstract.tex
\begin{abstract}

 Multimodal large language models (MLLMs) have rapidly advanced, yet their adoption in medicine remains limited by gaps in domain coverage, modality alignment, and grounded reasoning. In this work, we introduce \textbf{MedMO}, a medical foundation model built upon a generalized MLLM architecture and trained exclusively on large-scale, domain-specific data. MedMO follows a multi-stage training recipe: (i) cross-modal pretraining to align heterogeneous visual encoders with a medical language backbone; (ii) instruction tuning on multi-task supervision that spans captioning, VQA, report generation, retrieval, and grounded disease localization with bounding boxes; and (iii) reinforcement learning with verifiable rewards that combine factuality checks with a box-level GIoU reward to strengthen spatial grounding and step-by-step reasoning in complex clinical scenarios.
MedMO consistently outperforms strong open-source medical MLLMs across multiple modalities and tasks. \textbf{MedMO-8B-Next} leads all comparisons: on VQA benchmarks, it achieves an average improvement of \textbf{+6.6\%} over Fleming-VL-8B, with gains of \textbf{+6.0\%} on MMMU-Med, \textbf{+9.8\%} on PMC-VQA, and \textbf{+21.3\%} on MedXpertQA. For text-based QA, it attains \textbf{+14.4\%} over Fleming-VL-8B, driven by \textbf{+8.4\%} on MMLU-Med and \textbf{+30.1\%} on MedQA. In medical report generation, MedMO-8B-Next delivers \textbf{+6.7\%} on MIMIC-CXR. Moreover, it exhibits strong grounding capability with a Bacteria IoU of \textbf{56.1}, representing a \textbf{+47.8} IoU gain over Fleming-VL-8B, underscoring its robust spatial reasoning and localization performance. \textbf{MedMO-4B-Next} remains highly competitive at its smaller scale, surpassing Fleming-VL-8B across VQA, QA, and report generation benchmarks. Evaluations across radiology, ophthalmology, and pathology microscopy confirm MedMO's broad cross-modality generalization.

\end{abstract}

%% file: sec/1_intro.tex
\section{Introduction}
\label{sec:intro}
Recent advancements in Multimodal Large Language Models (MLLMs) have significantly accelerated progress across multimodal reasoning and understanding tasks~\citep{openai2024gpt4o,gemini2025gemini,chen2025januspro,kimi2025kimivl,zhu2025internvl3}. These models unify vision and language comprehension, achieving near-human performance on tasks such as image captioning, visual question answering (VQA), and multimodal reasoning. Despite their remarkable capabilities in general domains, their application to the medical domain remains substantially limited~\citep{yan2023multimodal,lee2023benefits,nori2023capabilities}. Biomedical data fundamentally differ from web-scale vision–language pairs: medical images demand precise, domain-specific interpretation, often requiring expert contextualization and robust grounding to textual clinical knowledge~\citep{li2025gmaivl}. As a result, general-purpose models frequently produce uncertain or hallucinated outputs when applied to medical tasks~\citep{li2023llavamed,chen2024huatuogptvision}.

To overcome these challenges, recent efforts have sought to adapt general-domain MLLMs into specialized medical multimodal models by incorporating domain-specific data and supervision~\citep{wu2023generalist,hyland2024maira1,zhang2024biomedgpt,zhang2024pmcvqa,seyfioglu2025quiltllava}. Early models such as LLaVA-Med\citep{li2023llavamed} leveraged PubMed-derived datasets for aligning medical images with textual knowledge, achieving foundational progress but limited by noisy data and narrow modality coverage. Subsequent works such as HuatuoGPT-Vision \citep{chen2024huatuogptvision}, GMAI-VL \citep{li2025gmaivl}, and HealthGPT \citep{lin2025healthgpt} introduced high-quality datasets, refined post-training strategies, and scaling recipes that improved alignment and reasoning. Parallel advancements in reasoning-based models, such as OpenAI’s o-series~\citep{openai2024openaio1,openai2025o3} and DeepSeek-R1~\citep{deepseekai2025deepseekr1}, as well as reinforcement learning with verifiable rewards (RLVR)\citep{shao2024deepseekmath,yu2025dapo}, have inspired recent medical research efforts\citep{lai2025medr1,pan2025medvlmr1} toward enhancing reasoning reliability and factual grounding in clinical scenarios.

Nevertheless, three key limitations persist in existing medical MLLMs.
\textbf{(1) }The majority rely on distilled data from advanced proprietary models~\citep{openai2024gpt4o,openai2024openaio1,openai2025gpt41,openai2025o3,gemini2025gemini,google2025gemini25pro}, which, while scalable, often lack accurate domain grounding, particularly for fine-grained clinical reasoning.
\textbf{(2)} Distillation pipelines frequently depend solely on generative outputs without structured supervision, amplifying hallucination risks and inconsistencies.
\textbf{(3)} Current models focus on individual tasks or narrow modality subsets (e.g., radiology or pathology) rather than achieving unified, cross-modal generalization across the diverse imaging modalities prevalent in real-world healthcare.

To bridge these gaps, we introduce MedMO, a powerful open-source post-trained multimodal large vision–language model (VLM) purpose-built for comprehensive medical image understanding and grounding (See Figure~\ref{fig:intro_fig}). MedMO is developed through a scalable and modular post-training pipeline, emphasizing progressive multimodal alignment, domain-specific reasoning, and cross-modal robustness.
We curate and harmonize a 26M+ with 45 open-source multimodal dataset, combining diverse medical imaging modalities (radiology, pathology, ophthalmology, dermatology, CT, MRI, ultrasound, and surgical videos) with carefully aligned text sources from open biomedical corpora and general-domain visual data.
Through multi-stage post-training, MedMO progressively enhances its capacity for visual grounding, clinical reasoning, and textual alignment, establishing a scalable pipeline toward a generalist foundation multimodal model for medical AI.

We further conduct comprehensive experiments and analyses on data curation, training, and alignment strategies, providing a transparent and reproducible framework for future medical MLLM development.
Extensive evaluations demonstrate that MedMO achieves state-of-the-art (SOTA) performance across diverse benchmarks, surpassing prior open and proprietary systems on tasks including medical VQA, report generation, and diagnostic reasoning.

Our main \textbf{contributions} are summarized as follows:
\begin{itemize}
\item We develop a powerful open-source post-trained multimodal large VLM, \textbf{MedMO}, designed for comprehensive medical image understanding and grounding.

\item We curate over 26M multimodal medical and biomedical samples from \textbf{45 datasets} and establish a multi-stage post-training that progressively enhances cross-modal alignment and reasoning. This provides a scalable roadmap toward a generalist foundation model for medical.
\item To evaluate VLM performance on detection tasks, we construct a dedicated Cell dataset from opensource microscopy images with varying sizes, shapes, and densities.
\item We conduct extensive experiments and analyses across data and methodology dimensions, providing an open benchmark for future multimodal medical LLM research and training recipes.
\end{itemize}

\section{Related Works}
\label{sec:related_works}
\subsection{Medical Language Multi-model Models} 
The rapid progress of LLMs has catalyzed remarkable advances in medical images. Building upon the success of general-domain LLMs, researchers have developed domain-adapted medical MLLMs that integrate visual and textual reasoning for clinical understanding~\citep{tian2023role,alSaad2024multimodal}.
Early efforts such as LLaVA-Med~\citep{li2023llavamed}, Med-Flamingo~\citep{moor2023medflamingo}, Qilin-MedVL~\citep{liu2023qilinmedvl}, and BioMedGPT~\citep{zhang2024biomedgpt} established the first medical vision–language models by aligning specialized visual encoders with pre-trained LLMs via linear projection layers, enabling foundational multimodal reasoning. However, these early systems were constrained by limited data diversity and suboptimal modality alignment, leading to hallucinations and factual inconsistencies~\citep{li2023llavamed,chen2024huatuogptvision}. Subsequent studies expanded this paradigm through richer datasets~\citep{ikezogwo2023quilt1m,li2025gmaivl,hamamci2025ctrate}, improved training strategies~\citep{nath2025vilam3,wang2024chatcad}, efficient fine-tuning~\citep{lin2025healthgpt}, and reinforcement learning~\citep{lai2025medr1,pan2025medvlmr1}. Proprietary systems such as Med-Gemini\citep{google2024medgemini} and Med-PaLM~\citep{singhal2023large,singhal2023expert} have further integrated multimodal and structured data for advanced reasoning, achieving strong performance across diagnostic and question-answering tasks~\citep{xie2024preliminary,saab2024capabilities,yang2024advancing,aydin2025openai,arora2025healthbench}.
Concurrently, specialized MLLMs targeting specific clinical contexts such as pathology~\citep{lu2024pathchat,wang2024pathology,zhao2024biomedparse,seyfioglu2025quiltllava}, radiology~\citep{hyland2024maira1,christensen2024echoclip,shui2025large,chaves2025llavarad,pai2025ctfm,tanno2024flamingocxr}, and ophthalmology~\citep{deng2024ophglm} have emerged, highlighting the growing demand for fine-grained, modality-aware intelligence in medical. Recent SOTA frameworks, such as Lingshu~\cite{xu2025lingshu} and Fleming-vl~\cite{shu2025fleming}, have improved the integration of medical and natural VLM tasks. However, their capabilities remain limited to selective tasks. Building on these foundations, our work emphasizes large-scale open-source post-training and progressive multimodal alignment. MedMO adopts a multi-stage design leveraging over 26M diverse multimodal samples, unifying heterogeneous medical modalities and textual data to achieve substantial gains across diverse clinical tasks.
\subsection{Grounding using multi-model models}
Unlike detection objective-based approaches such as grounding-DINO~\cite{liu2024grounding}, recent flagship VLMs have moved beyond captioning/VQA to explicit visual grounding~\cite{liu2024grounding,chen2025internvl25,gemini2025gemini} as well as point grounding~\cite{deitke2024pixmo}, i.e., returning spatial evidence such as bounding boxes or points aligned to textual queries. The Qwen2.5-VL~\cite{bai2025qwen25vl} report highlights grounding as a built-in capability, emphasizing precise object localization and event localization in long videos through native dynamic-resolution processing and absolute time encoding. Qwen2.5-VL generates grounded outputs in JSON with absolute coordinates, supporting both boxes and point clicks~\cite{deitke2024pixmo}.
Although the technical report is general-domain, these grounding primitives transfer to clinical data. For instance, MedSG-Bench~\cite{yue2025medsg} evaluates sequential medical grounding (difference/consistency grounding across image series) and explicitly benchmarks Qwen2.5-VL alongside medical-domain MLLMs (e.g., HuatuoGPT-Vision~\cite{chen2024huatuogpt}), finding that even advanced VLMs still face challenges on fine-grained, clinically realistic localization tasks-underscoring the need for domain-aligned post-training.

%% file: sec/2_methodology.tex
\section{Methodology-MedMO}
\label{sec:methodology}

\begin{figure*}
    \centering
    \includegraphics[width=0.95\linewidth]{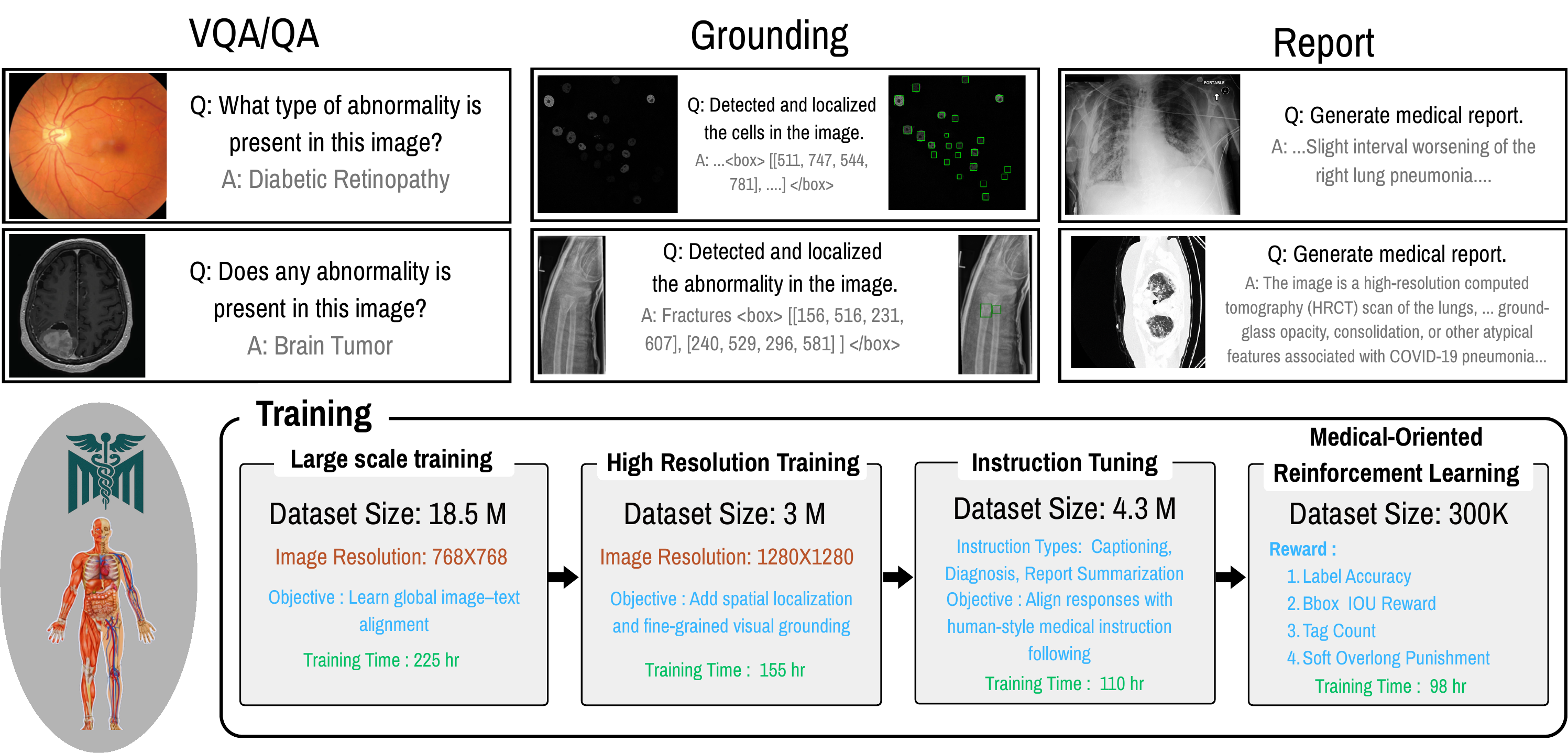}
    \caption{\textbf{Overview of the multi-stage training pipeline for medical image analysis.} The workflow consists of three main capabilities: (Top row) VQA/QA for identifying abnormalities in medical images, Grounding for spatial localization of detected features with bounding box coordinates, and Report generation for producing detailed medical reports. (Bottom) The training pipeline progresses through four sequential stages: (1) \textit{Large-scale training} on 18.5M image-text pairs at 768$\times$768 resolution for global image-text alignment, (2) \textit{High-resolution training} on 3M samples at 1280$\times$1280 resolution to enhance spatial localization and fine-grained visual grounding, (3) \textit{Instruction tuning} on 4.3M samples covering captioning, diagnosis, and report summarization tasks to align responses with human-style medical instruction following, and (4) \textit{Medical-oriented reinforcement learning} on 300K samples optimized using four reward signals: label accuracy, bounding box IoU, tag count, and soft overlap punishment. The complete pipeline for the MedMO-8B.}
    \label{fig:main_fig}
\end{figure*}

The overall methodology and multi-stage training pipeline are provided in Figure~\ref{fig:main_fig}. Starting from the Qwen3-VL-8B-Instruct model\footnote{\href{https://huggingface.co/Qwen/Qwen3-VL-8B-Instruct}{Qwen/Qwen3-VL-8B-Instruct}}, our approach consists of four sequential post-training stages: (1) General SFT aimed to train on large-scale instruction data to build foundational medical understanding; (2) High-quality medical image supervised fine-tuning, focused on expert-curated data to enhance visual grounding; (3) Instruction tuning and grounding fine-tuning, which align the model with clinical answering and spatial localization tasks; and (4) Reinforcement learning, designed to further improve instruction-following behavior and grounding accuracy. The following subsections provide an overview of the supervised fine-tuning strategy, followed by detailed descriptions of each stage.
\subsection{Overview of Supervised Fine-tuning}
\label{subsec:overview}

Our supervised fine-tuning (SFT) approach follows the standard next-token prediction paradigm for vision-language models. Given a multimodal input consisting of an image $\mathbf{v}$ and text sequence $\mathbf{x} = \{x_1, x_2, \ldots, x_n\}$, the model learns to predict the target response $\mathbf{y} = \{y_1, y_2, \ldots, y_m\}$ by maximizing the conditional likelihood:

\begin{equation}
\mathcal{L}_{\text{SFT}} = -\sum_{i=1}^{m} \log p_\theta(y_i \mid \mathbf{v}, \mathbf{x}, y_{<i}),
\end{equation}

\noindent where $\theta$ represents the model parameters, and $y_{<i}$ denotes all previously generated tokens.
MedMO builds upon the Qwen3-VL architecture, which consists of three primary components: (1) a vision encoder $\mathcal{E}_v$ that processes input images into visual representations; (2) a vision–language adapter $\mathcal{A}$ that projects multi-level ViT features into the language model’s embedding space through a DeepStack fusion mechanism, capturing fine-grained visual details and enhancing image–text alignment; and (3) a large language model decoder $\mathcal{D}$ that generates textual responses.

\subsection{Stage 1: General Medical SFT}
\label{subsec:stage1}

The first stage aims to establish foundational medical knowledge across diverse modalities and clinical scenarios. We utilize the publicly available \textbf{MedTrinity} dataset~\cite{xie2025medtrinity}, comprising \textbf{18.5M} large-scale instruction-following samples. This dataset $\mathcal{D}_{\text{general}}$ spans multiple imaging modalities (X-ray, CT, MRI, ultrasound, pathology, etc.) and includes captioning, visual question answering (VQA), and general-domain multimodal tasks, as illustrated in Figure~\ref{fig:dataset_modalaties}.

\noindent The Stage 1 dataset consists of:
\begin{itemize}[noitemsep,topsep=2pt,leftmargin=1.5em]
    \item \textbf{Medical image captioning:} $\mathcal{D}_{\text{caption}}$ with detailed textual descriptions of medical images.
    \item \textbf{Medical VQA:} $\mathcal{D}_{\text{vqa}}$ covering disease identification, anatomical recognition, and reasoning tasks.
    \item \textbf{General multimodal data:} $\mathcal{D}_{\text{general-mm}}$ for maintaining broad visual–language alignment.
\end{itemize}
\noindent The combined dataset is defined as:
\begin{equation}
\mathcal{D}_{\text{stage1}} = 
\mathcal{D}_{\text{caption}} \cup 
\mathcal{D}_{\text{vqa}} \cup 
\mathcal{D}_{\text{general-mm}}.
\label{eq:dataset}
\end{equation}

\subsection{Stage 2: Quality Medical Image and Grounding}
\label{subsec:stage2}

The second stage of SFT focuses on high-quality, expert-annotated medical image–text pairs to strengthen visual understanding and introduce grounding capability. We curate a refined dataset $\mathcal{D}_{\text{hq}}$ that includes both standard image–text supervision and medical grounding datasets containing bounding-box annotations (e.g., Chest X-ray, Wrist X-ray, Cell Microscopy, and CT). This stage extends the model’s visual encoder to predict localized features and bounding box coordinates, enabling spatial awareness while preserving global image–text alignment. Training objectives remain consistent with Stage~\ref{subsec:stage1}, combining captioning and VQA with supervised grounding signals.\\
\noindent\textbf{Grounding Dataset.} The grounding dataset $\mathcal{D}_{\text{ground}}$ includes: (1) Object detection annotations for anatomical structures and lesions, (2) Referring expression comprehension, and (3) Visual grounding QA pairsfor spatial localization.

\subsection{Stage 3: Instruction Tuning}
\label{subsec:stage3}

The third stage aligns MedMO’s responses with human-style medical reasoning through instruction tuning. Using a dataset $\mathcal{D}_{\text{inst}}$ of 4.3M multimodal instruction–response pairs, this phase covers captioning, diagnostic question answering, report summarization, and retrieval-based reasoning tasks. Instruction tuning improves task generalization and factual consistency, integrating clinical context understanding into both text- and vision-guided reasoning.
\subsection{Stage 4: Reinforcement Learning}
\label{subsec:stage4}
The final stage employs GRPO~\citep{shao2024deepseekmath} to enhance instruction-following capabilities through preference learning.\\
\noindent \textbf{GRPO Objective.} It optimizes the model by comparing multiple sampled responses for the same input. For each input $(\mathbf{v}, \mathbf{x})$, we sample $G$ responses $\{\mathbf{y}^{(1)}, \ldots, \mathbf{y}^{(G)}\}$ from the current policy $\pi_\theta$. Each response is evaluated using a reward function $r(\mathbf{v}, \mathbf{x}, \mathbf{y})$ that measures quality. \\
We follow the same objective as in GRPO~\cite{deepseekai2025deepseekr1,shao2024deepseekmath} with clip-higher and token level loss motivated from from DAPO~\cite{yu2025dapo}. For $(q,a)\sim\mathcal{D},~\{o_i\}_{i=1}^G\sim\pi_{\theta_{\text{old}}}(\cdot|q)$,
\vspace{-10pt}
\begin{align}
J(\theta) = & \mathbb{E}_{(q,a),o_i} \Bigg[\frac{1}{\sum_{i=1}^G |o_i|} \sum_{i=1}^G |o_i| \sum_{t=1} \min\Big( r_{i,t}(\theta) \hat{A}_{i,t}, \notag \\
& \quad \text{clip}\big(r_{i,t}(\theta), 1 - \varepsilon_{\text{low}}, 1 + \varepsilon_{\text{high}}\big) \hat{A}_{i,t} \Big) \Bigg]
\end{align}
\begin{align}
r_{i,t}(\theta) &= \frac{\pi_\theta(o_{i,t} \mid q, o_{i,<t})}{\pi_{\theta_{\text{old}}}(o_{i,t} \mid q, o_{i,<t})}, \label{eq:ratio} \\
\hat{A}_{i,t} &= \frac{R_i - \text{mean}(\{R_i\}_{i=1}^G)}{\text{std}(\{R_i\}_{i=1}^G)}. \label{eq:advantage}
\end{align}

\noindent The KL divergence term ensures the policy doesn't deviate too far from the reference model $\pi_{\text{ref}}$:

\begin{equation}
\mathcal{L}_{\text{KL}} = \mathbb{E}_{(\mathbf{v},\mathbf{x},\mathbf{y})} \left[ D_{\text{KL}}(\pi_\theta(\cdot \mid \mathbf{v}, \mathbf{x}) \| \pi_{\text{ref}}(\cdot \mid \mathbf{v}, \mathbf{x})) \right].
\end{equation}
For the reward function, we combine label accuracy, bounding-box reward, tag count, and soft-overlap penalty (see Fig.~\ref{fig:main_fig}). While these components are common in RL-based training, we introduce the Bounding Box Reward as a verifiable, spatially grounded signal that directly enhances localization performance.
\subsubsection{Bounding Box Reward}
\label{subsubsec:bbox_reward}

Given ground truth boxes $\mathcal{G}=\{g_j\}_{j=1}^{G}$ and predictions $\mathcal{P}=\{p_i\}_{i=1}^{P}$ in XYXY format and $\mathrm{GIoU}_{ij}\in[-1,1]$~\cite{rezatofighi2019generalized}, we score pairs via
\[
L1_{ij}=\frac{|x^p_1-x^g_1|+|y^p_1-y^g_1|+|x^p_2-x^g_2|+|y^p_2-y^g_2|}{2\sqrt{H^2 + W^2}}
\]
Normalize by the average image dimension makes the denominator resolution-invariant and proportional to image diagonal length.
We obtain a one-to-one assignment $M\subseteq\{1\ldots P\}\times\{1\ldots G\}$ by Hungarian matching on
\[
C_{ij}=w^m_{\text{L1}}\,L1_{ij}+w^m_{\text{G}}\,(1-\mathrm{GIoU}_{ij}),\quad
w^m_{\text{L1}}=5,\; w^m_{\text{G}}=2.
\]
For each matched pair $(i,j)\in M$, define a per-pair quality
\[
s_{ij}=\frac{w_{\text{L1}}\,(1-\mathrm{clip}_{[0,1]}(L1_{ij})) + w_{\text{G}}\left(\frac{\mathrm{GIoU}_{ij}+1}{2}\right)}{w_{\text{L1}}+w_{\text{G}}},
\]
where $w_{\text{L1}}=5,\; w_{\text{G}}=2$. The reward is a coverage-normalized sum with optional FP/FN penalties (Pen):
\[
\text{B}=\frac{1}{G}\sum_{(i,j)\in M} s_{ij},~
\text{Pen}=\frac{\lambda_{\text{FN}}(G-|M|)+\lambda_{\text{FP}}(P-|M|)}{\max(1,G)},
\]
\[
\boxed{R_{\text{bbox}}=\mathrm{clip}_{[0,1]}\big(\text{B}-\text{Pen}\big)}\footref{fn:supp}.
\]

%% file: sec/3_experiments.tex
\section{Experiments}\label{sec:expriments}
\subsection{Experimental Setup}\label{subsec:exp_setup}

MedMO was trained using \textbf{64$\times$ AMD Instinct MI210 GPUs} (64\,GB each) for \textbf{25 days} following a four-stage progressive pipeline (Figure~\ref{fig:main_fig}). The stages comprised: large-scale general medical SFT on 18.5M image–text pairs at 768$\times$768 resolution (\textbf{225\,h}); high-resolution fine-tuning on 3M curated samples at 1280$\times$1280 (\textbf{155\,h}); instruction tuning on 4.3M multimodal examples covering captioning, diagnosis, and report summarization (\textbf{110\,h}); and medical-oriented reinforcement learning on 300K samples with rewards for label accuracy, bounding-box IoU (\textbf{98\,h}). We follow standard VLM training practices using TRL~\cite{vonwerra2022trl}. Stage 1 uses BS = 10, LR = 1e-5, cosine schedule, and grad accum = 2. Stage 2 adopts BS = 2, LR = 8e-6, cosine schedule, and grad accum = 8. Stage 3 employs BS = 10 and LR = 5e-6 with grad accum = 2 for stable convergence \footnote{\label{fn:supp}For more details, please see our Supplementary Material.}.

\subsection{Datasets}
\begin{figure}[h!]
    \centering
    \includegraphics[width=0.9\linewidth]{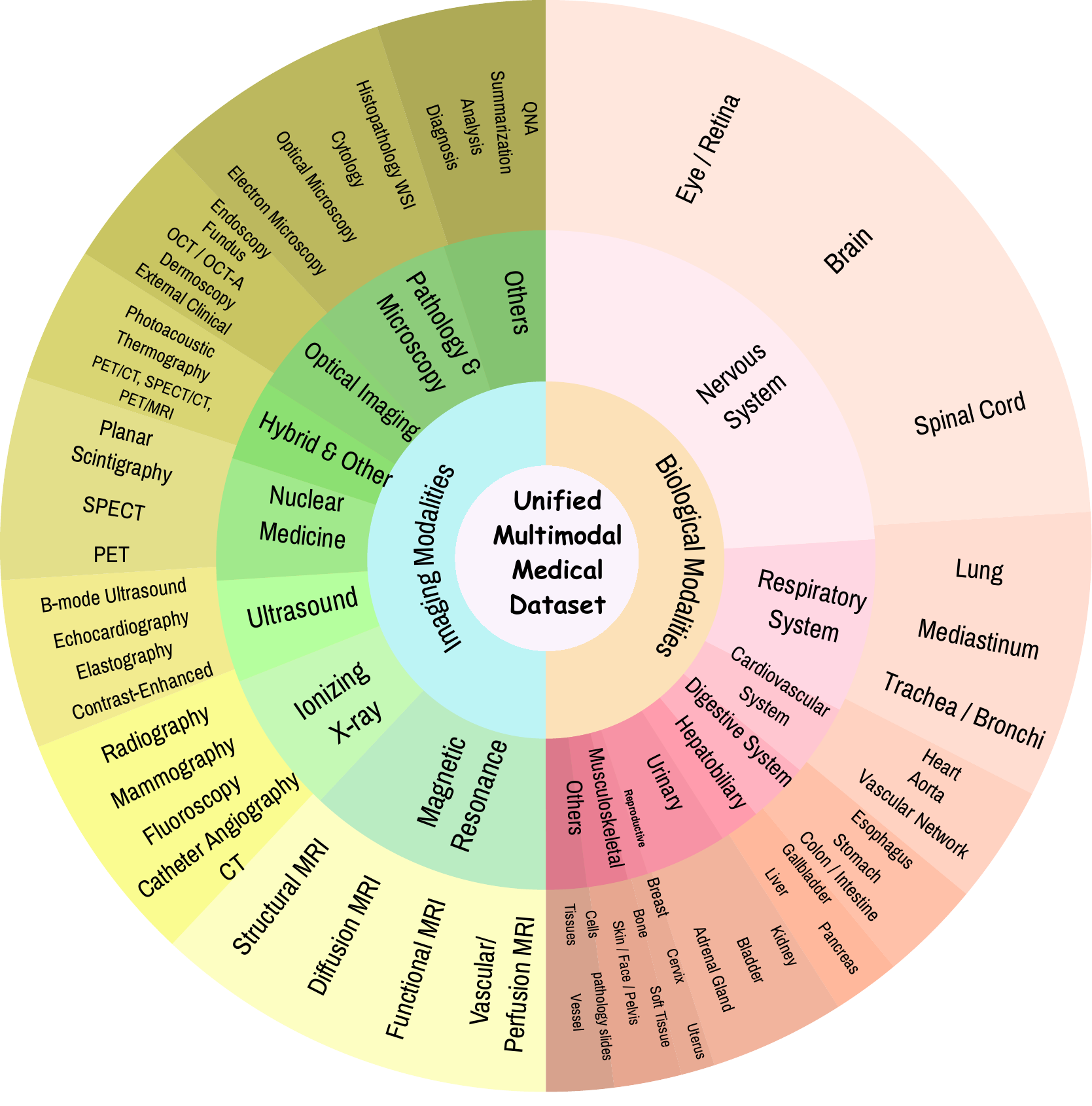}
    \caption{Composition of the unified multi-modal medical dataset comprising diverse imaging modalities and biological systems. }
    \label{fig:dataset_modalaties}
\end{figure}

We assembled a unified multimodal corpus of \textbf{45 datasets} spanning radiology, pathology, ophthalmology, dermatology, and surgical imaging, totaling over \textbf{26M samples}. The \textbf{MedTrinity} dataset~\cite{xie2025medtrinity} forms the core, contributing \textbf{18.5M} public instruction-following pairs. The corpus combines image–text and text-only data across diverse medical domains and clinical tasks. The dataset (Figure~\ref{fig:dataset_modalaties}) covers both \emph{imaging modalities} (e.g., X-ray, CT, MRI, ultrasound, optical, and nuclear imaging) and \emph{biological systems} (chest, brain, heart, liver, kidney, eye, colon, and tissue). For grounding tasks, we additionally used datasets with bounding-box annotations, including \emph{Chest X-ray, Wrist X-ray, Cell microscopy,} and \emph{CT} images. This comprehensive coverage supports robust multimodal understanding, spatial reasoning, and medical grounding. We curate a Cell Benchmark Dataset from open-source microscopy images\footref{fn:supp}, such as DeepCell~\cite{bannon2021deepcell} and Bacteria~\cite{van2018spatially}, covering diverse cell counts and densities\footref{fn:supp}.

\begin{table*}[h!]
\centering
\caption{
Performance comparison across medical \textbf{VQA} and \textbf{Text QA} benchmarks.
\textbf{Bold} and \underline{underline} indicate the best and second-best results, respectively.
OMIVQA and MedXQA refer to the OmniMedVQA and MedXpertQA benchmarks.
}
\label{tab:merged_vqa_textqa}
\resizebox{\textwidth}{!}{%
\begin{tabular}{l|cccccccc|cccccccc}
\toprule
& \multicolumn{8}{c|}{\textbf{VQA Benchmarks}} & \multicolumn{8}{c}{\textbf{Text QA Benchmarks}} \\
\cmidrule(lr){2-9} \cmidrule(lr){10-17}
\textbf{Models}
& \textbf{MMMU-Med}
& \textbf{\makecell{VQA-RAD\\(closed/all)}}
& \textbf{\makecell{SLAKE\\(closed/all)}}
& \textbf{\makecell{PathVQA\\(all)}}
& \textbf{PMC-VQA}
& \textbf{OMVQA}
& \textbf{MedXQA}
& \textbf{Avg.}
& \textbf{MMLU-Med}
& \textbf{PubMedQA}
& \textbf{MedMCQA}
& \textbf{MedQA}
& \textbf{\makecell{Medbullets\\(op4/op5)}}
& \textbf{MedXQA}
& \textbf{SGPQA}
& \textbf{Avg.} \\
\midrule
\multicolumn{17}{c}{\textit{Closed-source Models}} \\
\midrule
GPT-4.1
  & 75.2 & 65.0 & 72.2 & 55.5 & 55.2 & 75.5 & 45.2 & 63.4
  & 89.6 & 75.6 & 77.7 & 89.1 & 77.0 & 30.9 & 49.9 & 70.0 \\
Claude Sonnet 4
  & 74.6 & 67.6 & 70.6 & 54.2 & 54.4 & 65.5 & 43.3 & 61.5
  & 91.3 & 78.6 & 79.3 & 92.1 & 80.2 & 33.6 & 56.3 & 73.1 \\
Gemini-2.5-Flash
  & 76.9 & 68.5 & 75.8 & 55.4 & 55.4 & 71.0 & 52.8 & 65.1
  & 84.2 & 73.8 & 73.6 & 91.2 & 77.6 & 35.6 & 53.3 & 69.9 \\
\midrule
\multicolumn{17}{c}{\textit{Open-source Models}} \\
\midrule
BiomedGPT
  & 24.9 & 16.6 & 13.6 & 11.3 & 27.6 & 27.9 & -- & --
  & -- & -- & -- & -- & -- & -- & -- & -- \\
Med-R1-2B
  & 34.8 & 39.0 & 54.5 & 15.3 & 47.4 & -- & 21.1 & --
  & 51.5 & 66.2 & 39.1 & 39.9 & 33.6 & 11.2 & 17.9 & 37.0 \\
MedVLM-R1-2B
  & 35.2 & 48.6 & 56.0 & 32.5 & 47.6 & 77.7 & 20.4 & 45.4
  & 51.8 & 66.4 & 39.7 & 42.3 & 33.8 & 11.8 & 19.1 & 37.8 \\
MedGemma-4B-IT
  & 43.7 & 72.5 & 76.4 & 48.8 & 49.9 & 69.8 & 22.3 & 54.8
  & 66.7 & 72.2 & 52.2 & 56.2 & 45.6 & 12.8 & 21.6 & 46.8 \\
LLaVA-Med-7B
  & 29.3 & 53.7 & 48.0 & 38.8 & 30.5 & 44.3 & 20.3 & 37.8
  & 50.6 & 26.4 & 39.4 & 42.0 & 34.4 & 9.9 & 16.1 & 31.3 \\
HuatuoGPT-V-7B
  & 47.3 & 67.0 & 67.8 & 48.0 & 53.3 & 74.2 & 21.6 & 54.2
  & 69.3 & 72.8 & 51.2 & 52.9 & 40.9 & 10.1 & 21.9 & 45.6 \\
BioMediX2-8B
  & 39.8 & 49.2 & 57.7 & 37.0 & 43.5 & 63.3 & 21.8 & 44.6
  & 68.6 & 75.2 & 52.9 & 58.9 & 45.9 & 13.4 & 25.2 & 48.6 \\
Qwen2.5VL-7B
  & 50.6 & 64.5 & 67.2 & 44.1 & 51.9 & 63.6 & 22.3 & 52.0
  & 73.4 & 76.4 & 52.6 & 57.3 & 42.1 & 12.8 & 26.3 & 48.7 \\
InternVL2.5-8B
  & 53.5 & 59.4 & 69.0 & 42.1 & 51.3 & 81.3 & 21.7 & 54.0
  & 74.2 & 76.4 & 52.4 & 53.7 & 42.4 & 11.6 & 26.1 & 48.1 \\
InternVL3-8B
  & 59.2 & 76.4/52.9 & 72.1/62.4 & 39.0 & 53.8 & 79.1 & 22.4 & 57.4
  & 77.5 & 75.4 & 57.7 & 62.1 & 50.2/42.8 & 13.1 & 31.2 & 51.2 \\
Lingshu-7B
  & 54.0 & 77.2/43.0 & 82.4/33.2 & 41.9 & 54.2 & 82.9 & 26.9 & 55.1
  & 69.6 & 75.8 & 56.3 & 63.5 & 62.0/53.8 & 16.4 & 27.5 & 53.1 \\
Fleming-VL-8B
  & 63.3 & 78.4/56.4 & \underline{86.9/80.0} & \underline{56.5} & 64.3 & 88.2 & 21.6 & 66.1
  & 71.8 & 74.0 & 51.8 & 53.7 & 40.5/37.3 & 12.1 & 24.9 & 45.7 \\
Qwen3VL-8B
  & 61.4 & 54.1/31.2 & 34.3/15.0 & 14.6 & 52.3 & 77.2 & 24.8 & 40.5
  & 79.3 & 70.4 & 60.0 & \underline{66.1} & 56.1/47.7 & 15.1 & 34.7 & 53.6 \\
\midrule

\rowcolor{green!20}MedMO-4B
  & 54.6 & 50.9/35.0 & 41.0/30.0 & 42.4 & 50.6 & 79.7 & 24.8 & 45.4
  & 75.7 & \underline{78.0} & 58.0 & 78.5 & 57.5/47.7 & 16.4 & 29.4 & 55.1 \\
\rowcolor{green!20}MedMO-4B-Next
  & 58.7 & 79.7/59.6 & 78.0/74.0 & \textbf{73.3} & \textbf{75.7} & \underline{90.6} & \underline{27.0} & \underline{68.5}
  & 74.8 & \textbf{78.2} & 58.1 & 78.3 & 57.4/47.6 & 16.5 & 29.5 & 55.0 \\
  
\rowcolor{green!20}MedMO-8B
  & \underline{64.6} & \underline{72.3/64.7} & 70.6/70.0 & 56.3 & 59.4 & 84.8 & 26.2 & 63.2
  & \textbf{81.0} & 77.6 & \textbf{65.0} & \textbf{84.3} & \textbf{66.5/60.2} & \underline{19.9} & \textbf{36.0} & \textbf{61.3} \\
\rowcolor{green!20}MedMO-8B-Next
  & \textbf{69.3} & \textbf{86.4/68.0} & \textbf{83.0/81.6} & 56.3 & \underline{74.1} & \textbf{93.3} & \textbf{42.9} & \textbf{72.7}
  & \underline{80.2} & \underline{75.6} & \underline{62.0} & \underline{83.8} & \underline{65.2/57.8} & \textbf{20.9} & \underline{35.5} & \underline{60.1} \\
\bottomrule
\end{tabular}%
}
\end{table*}

\begin{table*}[ht!]
\centering
\caption{Comparison of medical report generation performance on MIMIC-CXR, CheXpert Plus, IU-Xray, and Med-Trinity using semantic (ROUGE-L, CIDEr) and model-based (RaTE, Semb) metrics. Models highlighted in green denote our proposed MedMO, which achieves the best overall performance across all datasets.}
\label{tab:medical_report_generation}
\resizebox{\textwidth}{!}{%
\begin{tabular}{l|cccc|cccc|cccc|cccc}
\toprule
& \multicolumn{4}{c|}{\textbf{MIMIC-CXR}} 
& \multicolumn{4}{c|}{\textbf{CheXpert Plus}} 
& \multicolumn{4}{c|}{\textbf{IU-Xray}} 
& \multicolumn{4}{c}{\textbf{Med-Trinity}} \\
\cmidrule(lr){2-5} \cmidrule(lr){6-9} \cmidrule(lr){10-13} \cmidrule(lr){14-17}
\textbf{Models} 
& \textbf{ROUGE-L} & \textbf{CIDEr} & \textbf{RaTE} & \textbf{Semb} 
& \textbf{ROUGE-L} & \textbf{CIDEr} & \textbf{RaTE} & \textbf{Semb} 
& \textbf{ROUGE-L} & \textbf{CIDEr} & \textbf{RaTE} & \textbf{Semb} 
& \textbf{ROUGE-L} & \textbf{CIDEr} & \textbf{RaTE} & \textbf{Semb} \\
\midrule
\multicolumn{17}{c}{\textit{Closed-source Models}} \\
\midrule
GPT-4.1 & 9.0 & 82.8 & 51.3 & 23.9 & 24.5 & 78.8 & 45.5 & 23.2 & 30.2 & 124.6 & 51.3 & 47.5 & -- & -- & -- & -- \\
Claude Sonnet 4 & 20.0 & 56.6 & 45.6 & 19.7 & 22.0 & 59.5 & 43.5 & 18.9 & 25.4 & 88.3 & 55.4 & 41.0 & -- & -- & -- & -- \\
Gemini-2.5-Flash & 25.4 & 80.7 & 50.3 & 29.7 & 23.6 & 72.2 & 44.3 & 27.4 & 33.5 & 129.3 & 55.6 & 50.9 & -- & -- & -- & -- \\
\midrule
\multicolumn{17}{c}{\textit{Open-source Models}} \\
\midrule
Med-R1-2B & 19.3 & 35.4 & 40.6 & 14.8 & 18.6 & 37.1 & 38.5 & 17.8 & 16.1 & 38.3 & 41.4 & 12.5 & -- & -- & -- & -- \\
MedVLM-R1-2B & 20.3 & 40.1 & 41.6 & 14.2 & 20.9 & 43.5 & 38.9 & 15.5 & 22.7 & 61.1 & 46.1 & 22.7 & -- & -- & -- & -- \\
MedGemma-4B-IT & 25.6 & 81.0 & 52.4 & 29.2 & \textbf{27.1} & 79.0 & 47.2 & 29.3 & 30.8 & 103.6 & 57.0 & 46.8 & -- & -- & -- & -- \\
LLaVA-Med-7B & 15.0 & 43.4 & 12.8 & 18.3 & 18.4 & 45.5 & 38.8 & 23.5 & 18.8 & 68.2 & 40.9 & 16.0 & -- & -- & -- & -- \\
HuatuoGPT-V-7B & 23.4 & 69.5 & 48.9 & 20.0 & 21.3 & 64.7 & 44.2 & 19.3 & 29.6 & 104.3 & 52.9 & 40.7 & -- & -- & -- & -- \\
BioMediX2-8B & 20.0 & 52.8 & 44.4 & 17.7 & 18.1 & 47.9 & 40.8 & 21.6 & 19.6 & 58.8 & 40.1 & 11.6 & -- & -- & -- & -- \\
Qwen2.5VL-7B & 24.1 & 63.7 & 47.0 & 18.4 & 22.2 & 62.0 & 41.0 & 17.2 & 26.5 & 78.1 & 48.4 & 36.3 & 23.5 & 81.5 & 44.9 & 38.3 \\
InternVL2.5-8B & 23.2 & 61.8 & 47.0 & 21.0 & 20.6 & 58.5 & 43.1 & 19.7 & 24.8 & 75.4 & 51.1 & 36.7 & 13.5 & 47.1 & 42.5 & 12.8 \\
InternVL3-8B & 22.9 & 66.2 & 48.2 & 21.5 & 20.9 & 65.4 & 44.3 & 25.2 & 22.9 & 76.2 & 51.2 & 31.3 & 12.9 & 46.6 & 42.2 & 3.7 \\
Lingshu-7B & 30.8 & 109.4 & 52.1 & 30.0 & \underline{26.5} & 79.0 & 45.4 & 26.8 & \underline{41.2} & \underline{180.7} & \underline{57.6} & \underline{48.4} & 16.0 & 74.5 & 44.4 & 24.0 \\
Fleming-VL-8B & \textbf{35.7} & 132.5 & 56.7 & 33.6 & 26.1 & 82.2 & 47.1 & 40.1 & \textbf{44.9} & \textbf{198.6} & \textbf{66.0} & \textbf{51.3} & 13.1 & 35.8 & 41.9 & 18.1 \\
Qwen3VL-8B & 25.1 & 77.9 & 50.3 & 33.4 & 21.9 & 67.4 & 44.4 & 37.9 & 25.0 & 91.44 & 52.5 & 42.9 & 20.2 & 69.9 & 45.9 & 33.6 \\
\midrule

\rowcolor{green!20}MedMO-4B & 26.0 & 92.6 & 49.8 & 31.6 & 15.1 & 62.3 & 36.6 & 34.2 & 26.6 & 94.0 & 42.1 & 41.3 & 22.5 & 152.6 & 47.8 & 34.3 \\
\rowcolor{green!20}MedMO-4B-Next & 28.3 & 96.7 & 52.0 & 34.3 & 23.5 & 74.5 & 42.6 & 38.7 & 38.0 & 147.8 & 62.0 & 49.4 & 26.3 & 183.8 & 49.5 & 38.6 \\

\rowcolor{green!20}MedMO-8B & 31.7 & \underline{140.0} & \underline{57.1} & \underline{50.0} & 23.6 & \underline{87.5} & \underline{47.3} & \underline{42.2} & 31.1 & 169.7 & 45.3 & 41.3 & \underline{37.0} & \underline{270.4} & \underline{53.0} & \underline{39.2} \\
\rowcolor{green!20}MedMO-8B-Next & \underline{32.6} & \textbf{143.4} & \textbf{57.7} & \textbf{51.5} & 25.7 & \textbf{88.3} & \textbf{48.1} & \textbf{43.8} & 31.8 & 171.9 & 56.0 & 43.1 & \textbf{38.5} & \textbf{272.1} & \textbf{53.8} & \textbf{40.7} \\

\bottomrule
\end{tabular}%
}
\end{table*}

\subsection{Results and Analysis}\label{subsec:results_analysis}

\begin{figure*}[h]
    \centering
    \includegraphics[width=\linewidth]{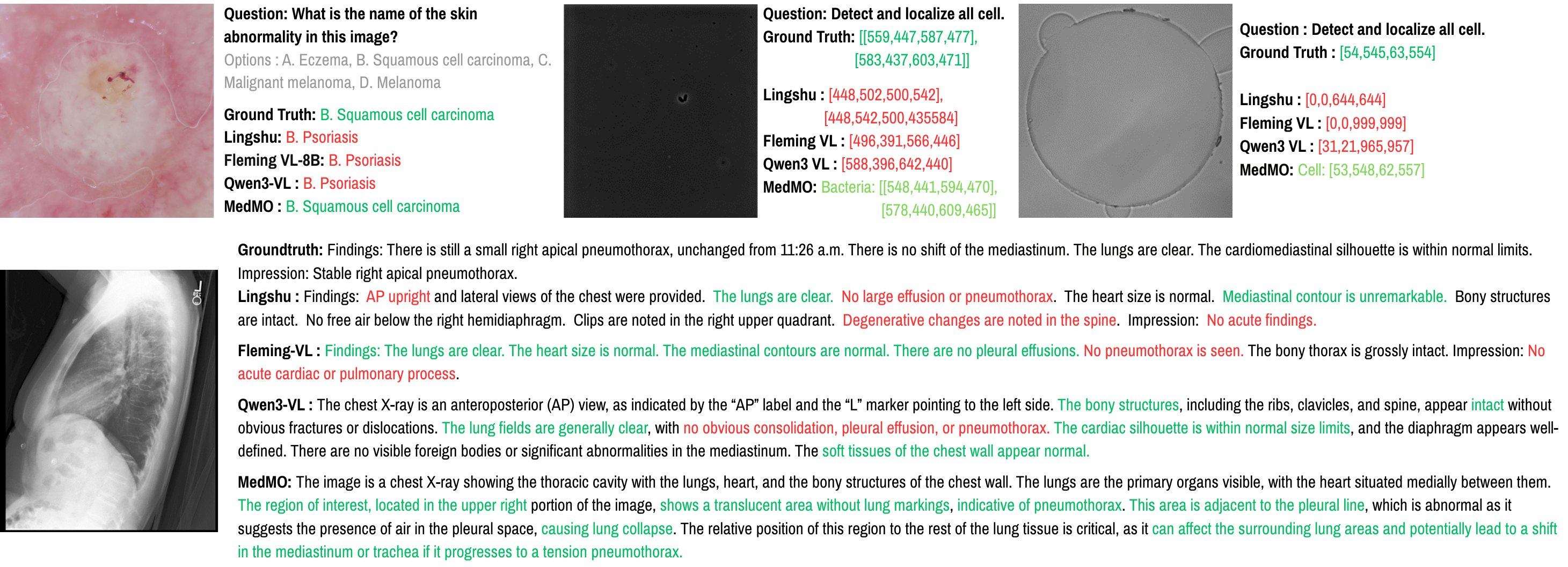}
    \caption{Qualitative comparison across diverse medical and visual question-answering tasks. Each block shows the ground truth, model predictions from Fleming-VL-8B (current Medical SOTA), Qwen3-VL (Baseline), and MedMO, and highlights textual or spatial alignment. MedMO provides more accurate medical understanding and localization in both diagnostic accuracy and clinical reasoning.}
    \label{fig:visual_analysis}
\end{figure*}

\subsubsection{SOTA Comparison of MedMO for QA}

Table~\ref{tab:merged_vqa_textqa} summarizes MedMO's performance across medical VQA and Text QA benchmarks for all four variants: MedMO-4B, MedMO-4B-Next, MedMO-8B, and MedMO-8B-Next.

\noindent\textbf{VQA Benchmarks.}
\textbf{MedMO-8B-Next} achieves the highest VQA average of \textbf{72.7\%}, outperforming all open-source competitors including Fleming-VL-8B (66.1\%) and Lingshu-7B (55.1\%) by \textbf{+6.6\%} and \textbf{+17.6\%}, respectively. It sets new state-of-the-art scores on MMMU-Med (\textbf{69.3\%}), VQA-RAD (\textbf{86.4/68.0}), SLAKE (\textbf{83.0/81.6}), and OMVQA (\textbf{93.3\%}). \textbf{MedMO-4B-Next} also surpasses Fleming-VL-8B with a VQA average of \textbf{68.5\%}, achieving competitive scores on PMC-VQA (\textbf{75.7\%}) and OMVQA (\textbf{90.6\%}) despite its smaller scale. The base variants MedMO-4B (45.4\%) and MedMO-8B (63.2\%) show consistent improvement with scale, with MedMO-8B notably achieving the second-best PathVQA score (\underline{56.3\%}).

\noindent\textbf{Text QA Benchmarks.}
\textbf{MedMO-8B-Next} achieves a Text QA average of \textbf{60.1\%}, outperforming Fleming-VL-8B (45.7\%) by \textbf{+14.4\%}. It leads on MMLU-Med (\textbf{80.2\%}), MedQA (\textbf{83.8\%}), and MedXpertQA (\underline{20.9\%}), demonstrating strong clinical reasoning and knowledge integration. \textbf{MedMO-8B} achieves the highest QA average among all models including \textit{Next} variants at \textbf{61.3\%}, leading on MedMCQA (\textbf{65.0\%}), MedQA (\textbf{84.3\%}), and Medbullets (\textbf{66.5/60.2}), suggesting its base instruction tuning yields strong reasoning without RL fine-tuning overhead. \textbf{MedMO-4B-Next} achieves a QA average of \textbf{55.0\%}, surpassing Fleming-VL-8B (45.7\%) by \textbf{+9.3\%} and even matching or exceeding Lingshu-7B (53.1\%) on several benchmarks including PubMedQA (\textbf{78.2\%}). Overall, all MedMO variants consistently outperform same-scale open-source models, with larger and \textit{Next} variants delivering substantial improvements across both VQA and QA tasks.

\subsubsection{SOTA Comparison of MedMO for Report Generation}

Table~\ref{tab:medical_report_generation} evaluates medical report generation across four datasets using semantic (ROUGE-L, CIDEr) and model-based (RaTE, Semb) metrics.

\noindent\textbf{MIMIC-CXR.}
\textbf{MedMO-8B-Next} achieves the highest CIDEr of \textbf{143.4} and strong RaTE (\textbf{57.7\%}) and Semb (\textbf{51.5\%}), outperforming Fleming-VL-8B (132.5, 56.7\%, 33.6\%) on all metrics except ROUGE-L, where Fleming leads (35.7\% vs.\ 32.6\%). \textbf{MedMO-8B} achieves the second-best CIDEr (\underline{140.0}) with the highest Semb among all models (\underline{50.0\%}), confirming that MedMO generates reports with superior semantic fidelity and clinical coherence. \textbf{MedMO-4B-Next} (CIDEr: 96.7, Semb: 34.3\%) and \textbf{MedMO-4B} (CIDEr: 92.6, Semb: 31.6\%) also outperform most open-source baselines despite their smaller scale.

\noindent\textbf{CheXpert Plus.}
\textbf{MedMO-8B-Next} achieves the highest CIDEr (\textbf{88.3}) and RaTE (\textbf{48.1\%}) and Semb (\textbf{43.8\%}), surpassing Fleming-VL-8B (82.2, 47.1\%, 40.1\%) across all model-based metrics. \textbf{MedMO-8B} achieves the second-best CIDEr (\underline{87.5}) and Semb (\underline{42.2\%}). While MedGemma-4B-IT leads on ROUGE-L (27.1\% vs.\ 25.7\%), MedMO's superior CIDEr and Semb scores indicate better semantic coherence and clinical accuracy over lexical overlap.

\noindent\textbf{IU-Xray.}
Fleming-VL-8B leads on IU-Xray with CIDEr 198.6, RaTE 66.0\%, and Semb 51.3\%. \textbf{MedMO-8B-Next} achieves competitive performance (CIDEr: 171.9, RaTE: 56.0\%, Semb: 43.1\%), and \textbf{MedMO-8B} ranks second on ROUGE-L (\underline{37.0\%}) and CIDEr (\underline{169.7\%}). \textbf{MedMO-4B-Next} shows a strong improvement over the base 4B variant, achieving CIDEr 147.8 and Semb 49.4\%, while Lingshu-7B leads on ROUGE-L (41.2\%) among open-source models.

\noindent\textbf{Med-Trinity.}
On Med-Trinity, which spans diverse modalities including CT, MRI, ultrasound, and pathology, \textbf{MedMO-8B-Next} achieves the highest ROUGE-L (\textbf{38.5\%}) and CIDEr (\textbf{272.1}), while \textbf{MedMO-8B} leads on RaTE (\textbf{53.0\%}) and Semb (\underline{39.2\%}). Both variants dramatically outperform all baselines — MedMO-8B-Next's CIDEr of 272.1 is over \textbf{3$\times$} higher than the next best open-source model, Qwen2.5VL-7B (81.5), underscoring MedMO's exceptional capability in multi-modal medical report generation. \textbf{MedMO-4B-Next} also delivers strong performance (CIDEr: 183.8), surpassing all non-MedMO baselines.


\begin{table}[ht!]
\centering
\caption{
Performance comparison of selected MLLMs on Medical Grounding Benchmarks.
NIH: Chest X-ray; DeepLesion: lesion detection; Bacteria: detection; 
MedSG: multi-view, object-tracking, and referring tasks. 
All values are IoU scores (\%), and ``Avg.'' denotes the mean across tasks.
}

\label{tab:segmentation_selected}
\resizebox{\columnwidth}{!}{%
\begin{tabular}{lccccccc}
\toprule
\textbf{Model} & \textbf{NIH} & \textbf{DeepLession} & \textbf{Bacteria} &
\textbf{\shortstack{MedSG \\ (multi\_view)}} &
\textbf{\shortstack{MedSG \\ (object\_tracking)}} &
\textbf{\shortstack{MedSG \\ (referring)}} &
\textbf{Avg.} \\
\midrule
InternVL3-8B   & 10.1  & 0.00 & 0.7  & 6.3 & 13.0 & 3.3  & 5.6 \\
Fleming-VL-8B  & 0.00  & 0.00 & 8.3  & 42.0 & 36.7 & 16.6 & 17.2 \\
Lingshu-7B     & 5.3   & 0.7  & 10.8 & 28.3 & 38.7 & 10.4 & 13.9 \\
Qwen3VL-8B     & \textbf{16.4}  & 0.00 & 9.16 & 8.4 & 17.8 & 31.4 & 13.8 \\
MedSG-Bench    & --    & --   & --   & 55.0 & 62.1 & 60.4 & - \\
\rowcolor{green!20}MedMO-8B & 8.83 & \underline{38.5} & \underline{54.6} & \underline{75.8} & \underline{77.2} & \underline{70.1} & \underline{54.2} \\
\rowcolor{green!20}MedMO-8B-Next & \underline{15.9} & \textbf{40.5} & \textbf{56.1} & \textbf{77.5} & \textbf{78.8} & \textbf{71.9} & \textbf{56.8} \\

\bottomrule
\end{tabular}%
}
\end{table}

\subsubsection{MedMO for Grounding}

Table~\ref{tab:segmentation_selected} reports IoU on six medical grounding tasks covering chest X-ray localization (NIH), lesion detection (DeepLesion), microscopy segmentation (Bacteria), and three MedSG subtasks (multi-view, object tracking, and referring expression grounding). \textbf{MedMO-8B-Next} achieves the best overall average at \textbf{56.8\%}, and \textbf{MedMO-8B} follows at \textbf{54.2\%}. Both results are substantially higher than the strongest baseline \textbf{Fleming-VL-8B} at \textbf{17.2\%} and \textbf{Lingshu-7B} at \textbf{13.9\%}.

On DeepLesion, MedMO-8B and MedMO-8B-Next reach 38.5\% and 40.5\% IoU, while Fleming-VL-8B, InternVL3-8B, and Qwen3VL-8B obtain 0.00\%. This contrast indicates that lesion localization is a major weakness for several existing medical vision language baselines and is a clear strength of MedMO. On Bacteria microscopy segmentation, MedMO-8B-Next achieves 56.1\% IoU and MedMO-8B achieves 54.6\%, which is far above the best competing baseline Lingshu-7B at 10.8\%. On NIH chest X-ray localization, MedMO-8B-Next reaches 15.9\%, which is close to the best score from Qwen3VL-8B at 16.4\%, and MedMO-8B improves over Fleming-VL-8B which scores 0.00\%.

MedMO also performs strongly on the MedSG benchmarks that test multi-view correspondence, temporal object tracking, and referring expression grounding. MedMO-8B-Next obtains 77.5\% on multi-view, 78.8\% on object tracking, and 71.9\% on referring expression grounding, while MedMO-8B achieves 75.8\%, 77.2\%, and 70.1\%. Both variants exceed the specialist MedSG-Bench scores of 55.0\%, 62.1\%, and 60.4\% on the three subtasks, and they also outperform the strongest general baselines such as Fleming-VL-8B (42.0\%, 36.7\%, 16.6\%) and Lingshu-7B (28.3\%, 38.7\%, 10.4\%). These results show consistent cross-task gains for MedMO on grounding and spatial reasoning across radiology, microscopy, and multi-task medical scene grounding.

\subsection{Ablation Study}
\label{subsec:ablation}

\subsubsection{Ablation on Post-Training Stages}
\begin{figure}[h]
    \centering
    \includegraphics[width=\linewidth]{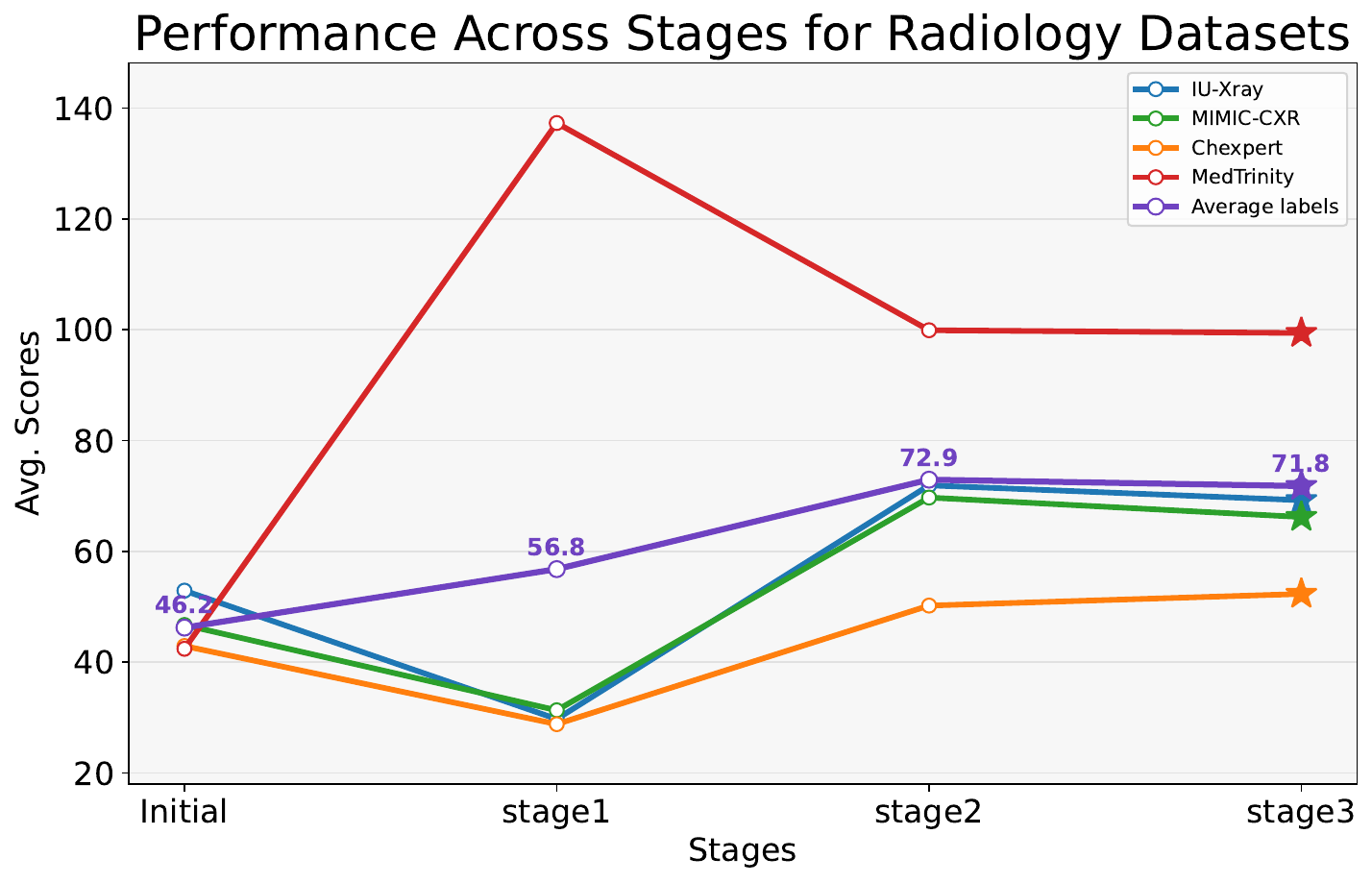}
    \caption{\textbf{Performance across post-training stages on radiology datasets.} MedMO exhibits consistent gains in diagnostic accuracy and localization across IU-Xray, MIMIC-CXR, CheXpert, and MedTrinity datasets. The sharp improvement at Stage~2 highlights the benefit of alignment tuning with medical reasoning objectives.}
    \label{fig:radiology_stage_performance}
\end{figure}

\begin{figure}[h]
    \centering
    \includegraphics[width=\linewidth]{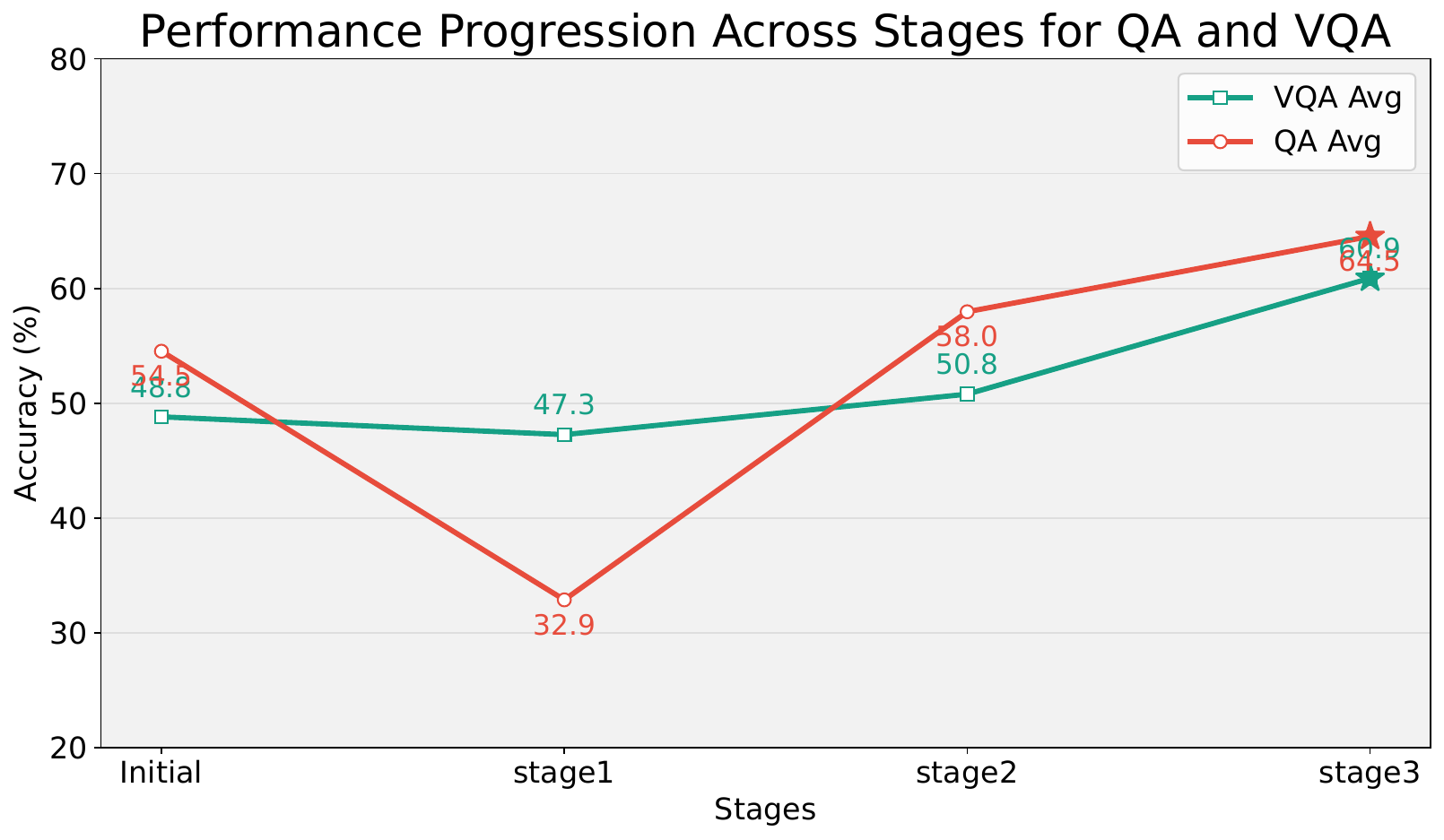}
    \caption{\textbf{QA and VQA ablation across stages.} Both question-answering (QA) and visual question-answering (VQA) accuracy improve progressively, demonstrating that stage-wise optimization enhances multimodal reasoning and factual grounding in medical.}
    \label{fig:qa_vqa_progress}
\end{figure}

We perform a stage-wise ablation to evaluate the contribution of each post-training phase to MedMO’s performance on radiology and QA benchmarks. 
As shown in Figures~\ref{fig:radiology_stage_performance} and~\ref{fig:qa_vqa_progress}, performance progressively improves across stages, validating the effectiveness of our optimization strategy. 
In Stage~1, the model trained on the MedTrinity dataset achieves strong accuracy on that dataset but shows slight degradation on others. 
Stage~2 provides the largest gain through high-resolution and diverse medical data training, while Stage~3 further boosts QA and VQA performance via instruction tuning, enhancing multimodal alignment and reasoning. 
Each stage contributes complementary improvements, leading to a consistent overall enhancement in MedMO’s performance across all tasks.

\subsubsection{Bounding-Box Verifiable Reward}
Table \ref{tab:before_after_delta} shows consistent gains after reinforcement learning, confirming the effectiveness of our bounding-box reward. Even with small improvements, the reward reliably enhances spatial precision and grounding consistency across datasets\footref{fn:supp}.

\begin{table}[ht!]
\centering
\caption{ Absolute change ($\Delta$) after bouding box verifiable reward\footref{fn:supp}.} 
\label{tab:before_after_delta}
\resizebox{0.7\columnwidth}{!}{
\begin{tabular}{lccc}
\toprule
\textbf{Dataset} & \textbf{Before (IoU)} & \textbf{After (IoU)} & \textbf{$\Delta$ (IoU)} \\
\midrule
NIH              & 8.8  & 13.3 & \textcolor{ForestGreen}{+4.5} \\
DeepLesion       & 38.5 & 38.9 & \textcolor{ForestGreen}{+0.4} \\
Bacteria   & 54.6 & 55.0 & \textcolor{ForestGreen}{+0.4} \\
\bottomrule
\end{tabular}
}
\end{table}

\section{Conclusion}


We introduced \textbf{MedMO}, a general-purpose medical multimodal foundation model that unifies visual grounding, clinical reasoning, and language understanding across diverse medical modalities. MedMO is trained with a scalable four-stage post-training pipeline that includes large-scale alignment, high-resolution fine-tuning, instruction tuning, and reinforcement learning with verifiable rewards. This design enables robust multimodal comprehension and precise spatial localization. Experimental results show substantial gains over strong open-source baselines across VQA, text QA, report generation, and grounding benchmarks, establishing \textbf{MedMO} as the \textit{best fully open-source medical multimodal foundation model} to date. As an open medical MLLM, MedMO provides a scalable path toward reliable and transparent medical vision language systems. Future work could explore strategies to better retain SFT knowledge within reinforcement learning frameworks.

\paragraph{Limitation.} 

MedMO’s stage-wise training introduces minor task-level performance shifts, as shown in Figures~\ref{fig:radiology_stage_performance} and~\ref{fig:qa_vqa_progress}, a typical behavior in large multimodal models due to \textit{catastrophic forgetting}~\cite{luo2025empirical}. 
Future work will focus on improving cross-task retention while expanding coverage across additional medical imaging modalities.

%% file: sec/X_suppl.tex
\clearpage
\setcounter{page}{1}
\maketitlesupplementary
\appendix
\setcounter{section}{0}

\section{Reward function details}
\label{sec:rationale}
\subsection{Bounding Box Reward Function}
\label{subsubsec:bbox_reward}

For grounding tasks in the reinforcement learning stage, we employ a specialized reward function that evaluates the quality of predicted bounding boxes against ground truth annotations. This reward is computed using Hungarian matching combined with geometric metrics.

\paragraph{Notation and Setup.} Given ground truth boxes $\mathcal{G} = \{g_j\}_{j=1}^{G}$ and predicted boxes $\mathcal{P} = \{p_i\}_{i=1}^{P}$ in XYXY format (i.e., $(x_1, y_1, x_2, y_2)$ coordinates), we first determine the image dimensions $(H, W)$ from the maximum extents of ground truth boxes if available, otherwise from predictions (with fallback to $(1, 1)$ if both are empty).

\paragraph{Pairwise Metrics.} For each pair of boxes $(p_i, g_j)$, we compute two geometric measures:

\textbf{Normalized L1 Distance:} The L1 distance over all four coordinates, normalized by the image perimeter:
\begin{equation}
L1_{ij} = \frac{|x^p_1 - x^g_1| + |y^p_1 - y^g_1| + |x^p_2 - x^g_2| + |y^p_2 - y^g_2|}{2\sqrt{H^2 + W^2}}
\end{equation}

\textbf{Generalized IoU (GIoU):} We compute $\mathrm{GIoU}_{ij} \in [-1, 1]$ following~ \citet{rezatofighi2019generalized}, which extends standard IoU to account for non-overlapping boxes.

\paragraph{Hungarian Matching.} To establish optimal correspondence between predictions and ground truth, we construct a cost matrix:
\begin{equation}
C_{ij} = w^m_{\text{L1}} \cdot L1_{ij} + w^m_{\text{G}} \cdot (1 - \mathrm{GIoU}_{ij}),
\end{equation}
where $w^m_{\text{L1}} = 5.0$ and $w^m_{\text{G}} = 2.0$ are matching cost weights. We apply the Hungarian algorithm to find the minimum-cost bipartite matching, yielding $m = \min(P, G)$ matched pairs $\{(i_k, j_k)\}_{k=1}^{m}$.

\paragraph{Per-Match Score.} For each matched pair $(i_k, j_k)$, we compute a quality score by:
\begin{enumerate}
    \item Mapping GIoU to $[0, 1]$: $\tilde{G}_k = \frac{\mathrm{GIoU}_{i_k j_k} + 1}{2}$
    \item Clamping L1 to $[0, 1]$: $\hat{L1}_k = \mathrm{clip}_{[0,1]}(L1_{i_k j_k})$
    \item Computing weighted blend:
\end{enumerate}
\begin{equation}
s_k = \frac{w_{\text{L1}} \cdot (1 - \hat{L1}_k) + w_{\text{G}} \cdot \tilde{G}_k}{w_{\text{L1}} + w_{\text{G}}}, \quad s_k \in [0, 1]
\end{equation}
where $w_{\text{L1}} = 5.0$ and $w_{\text{G}} = 2.0$ are pair score weights.

\paragraph{Final Reward Computation.} The base reward is the coverage-normalized sum of matched pair scores:
\begin{equation}
\text{base} = \frac{1}{G} \sum_{k=1}^{m} s_k
\end{equation}

We optionally apply penalties for false positives (FP) and false negatives (FN):
\begin{equation}
\text{penalty} = \frac{\lambda_{\text{FN}} \cdot (G - m) + \lambda_{\text{FP}} \cdot (P - m)}{\max(1, G)},
\end{equation}
where $\lambda_{\text{FN}}$ and $\lambda_{\text{FP}}$ are penalty coefficients (default: 0). The final bounding box reward is:
\begin{equation}
\boxed{R_{\text{bbox}} = \mathrm{clip}_{[0,1]}\left(\text{base} - \text{penalty}\right)}
\end{equation}

Expanding the base term:
\begin{equation}
\text{base} = \frac{1}{G} \sum_{k=1}^{m} \frac{w_{\text{L1}}(1 - L1_{i_k j_k}) + w_{\text{G}} \cdot \frac{\mathrm{GIoU}_{i_k j_k} + 1}{2}}{w_{\text{L1}} + w_{\text{G}}}
\end{equation}

\paragraph{Edge Cases.} The reward function handles special cases as follows:
\begin{itemize}
    \item \textbf{No ground truth boxes} ($G = 0$): $R_{\text{bbox}} = 0.5$ (neutral reward)
    \item \textbf{Ground truth present but no predictions} ($G > 0, P = 0$): $R_{\text{bbox}} = \mathrm{clip}_{[0,1]}(0 - \text{penalty})$, which equals $0.0$ with default penalties
    \item \textbf{Failed matching} (no feasible pairs): Treated as $m = 0$, where all ground truth boxes are unmatched and all predictions are false positives
\end{itemize}

This reward formulation encourages the model to produce accurate bounding box predictions through Hungarian-matched optimization of both localization (L1) and overlap quality (GIoU), while penalizing missing detections and spurious predictions.

\section{Experimental Details}

We conducted all experiments using the SFT\_Trainer and RL (GRPO) trainer frameworks. Unless otherwise noted, we used mixed‐precision training (dtype=bfloat16) on a cluster of $64 \times$ AMD Instinct MI210 GPUs. Random seeds, optimizer state, and scheduler configuration were logged for full reproducibility.

\subsection{Stage\,1: General SFT}

\subsection*{Parameters Details}
We provide detailed experimental settings in Table \ref{tab:stage1}, which we apply exclusively to training stage 1 MedMO.

\begin{table}[h!]
\centering
\begin{tabular}{ll}
\hline
\textbf{Parameter} & \textbf{Value} \\
\hline
Batch size & 10 \\
Gradient accumulation steps & 2 \\
Learning rate (initial) & $1 \times 10^{-5}$ \\
LR scheduler & Cosine decay \\
Number of epochs & 1 \\
Image resolution & $768 \times 768$ pixels \\
dtype & bfloat16 \\
\hline
\end{tabular}
\caption{Training parameter details for stage 1.}
\label{tab:stage1}
\end{table}
\subsection*{Training Dynamics}
During Stage~1, optimization converges rapidly: the loss drops from $\sim\!11$ to $<\!0.3$ within the first $\approx\!10$ steps, and entropy collapses from $\sim\!5.3$ to $\sim\!0.1$ over the same window, indicating quickly sharpened token distributions. Mean token accuracy rises steeply from $\sim\!0.6$ to $\sim\!0.95$ by step $\approx\!10$ and then plateaus with minor oscillations thereafter. These curves reflect stable optimization under the cosine schedule, fast fit to the instruction format, and no signs of late-stage instability during the single-epoch SFT. \emph{Unless noted, one plotted ``step'' corresponds to an aggregate over 100 mini-batches (logging interval = 100 batches).}

\begin{figure}[h]
  \centering
  \includegraphics[width=\linewidth]{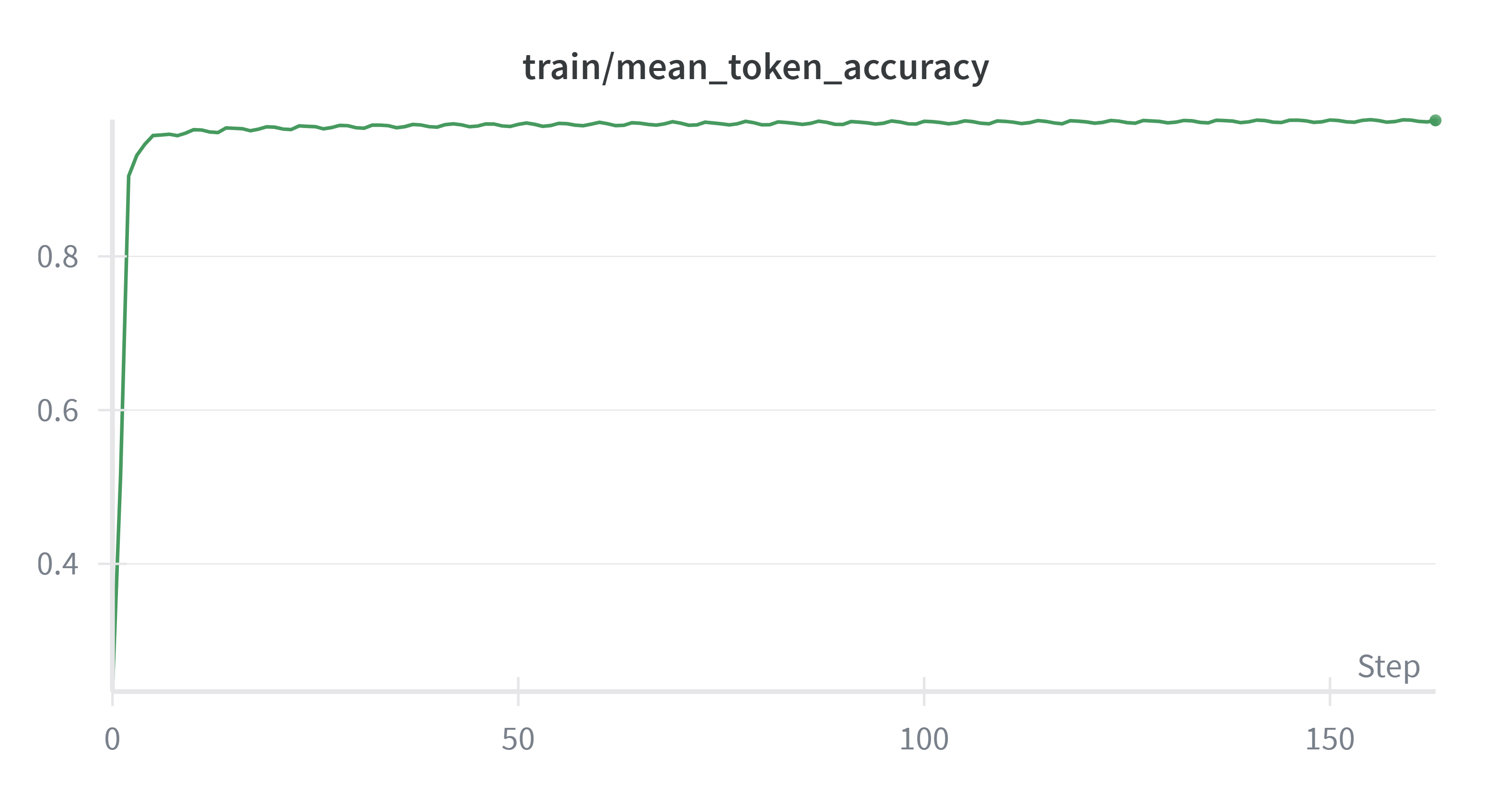}
  \caption{Stage~1 mean token accuracy vs.\ step (each step = 100 mini-batches). Accuracy jumps to $\sim$0.95 within $\approx$10 steps and remains stable.}
  \label{fig:stage1_acc}
\end{figure}

\begin{figure}[h]
  \centering
  \includegraphics[width=\linewidth]{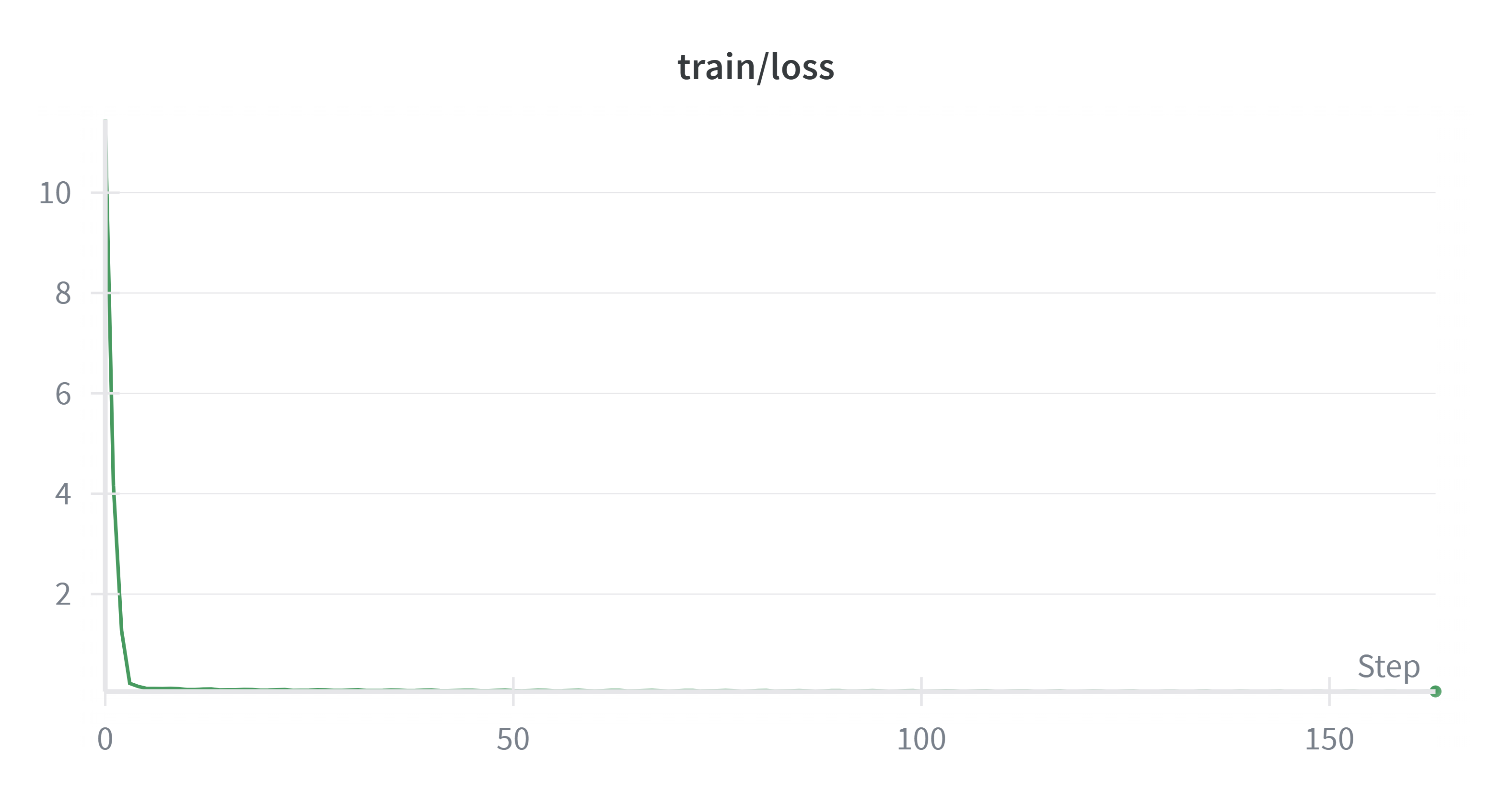}
  \caption{Stage~1 training loss vs.\ step (each step = 100 mini-batches). Loss declines from $\sim$11 to $<0.3$ in the first $\approx$10 steps, then flattens.}
  \label{fig:stage1_loss}
\end{figure}

\begin{figure}[h]
  \centering
  \includegraphics[width=\linewidth]{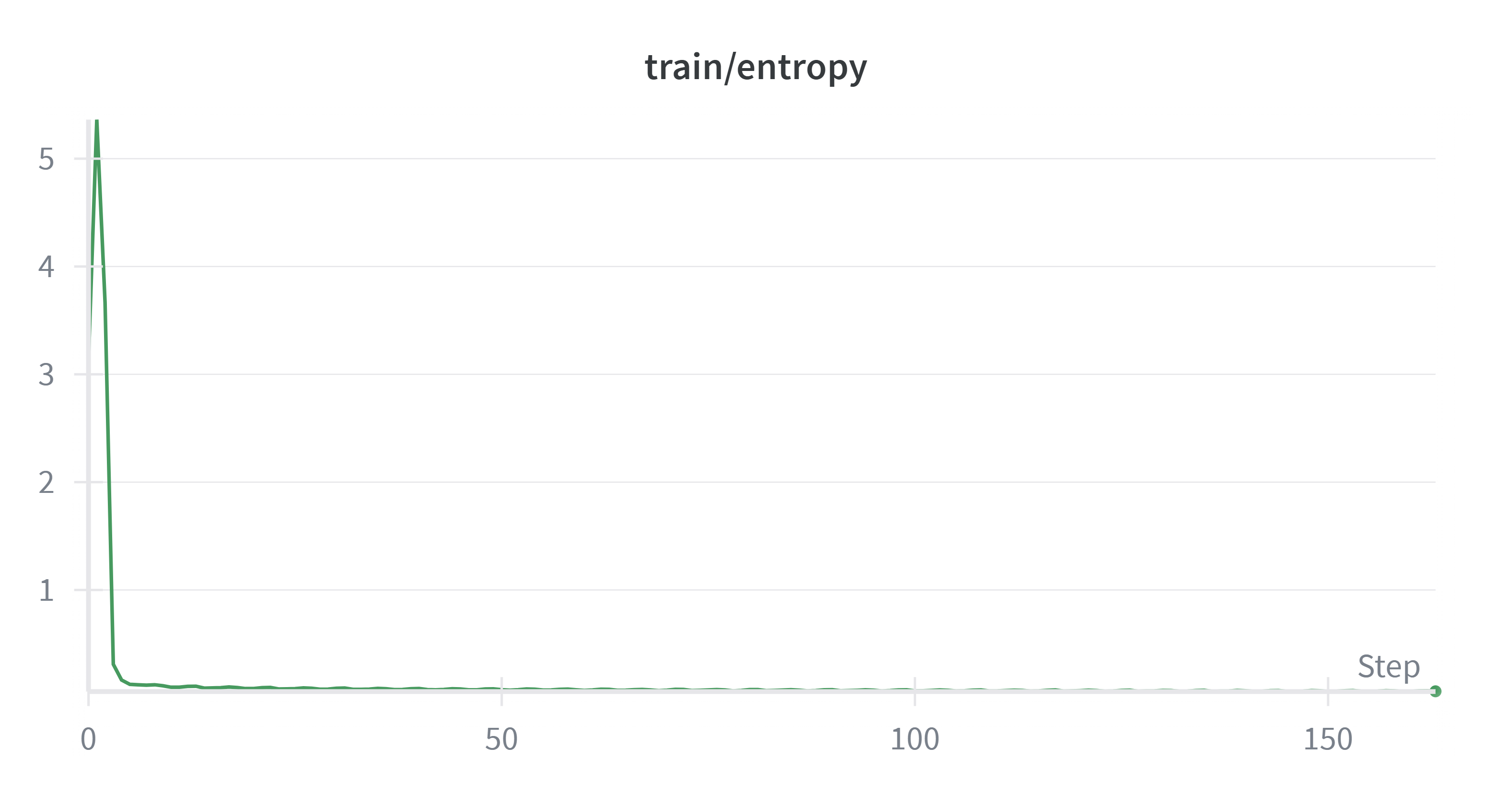}
  \caption{Stage~1 output entropy vs.\ step (each step = 100 mini-batches). Entropy collapses from $\sim$5.3 to $\sim$0.1 by $\approx$10 steps, indicating confident token distributions.}
  \label{fig:stage1_entropy}
\end{figure}

\subsection{Stage\,2: High-Resolution Image SFT}

\subsection*{Parameters Details}
We provide detailed experimental settings in Table \ref{tab:stage2_hpy}, which we apply exclusively to training stage 2 MedMO.
\begin{table}[h!]
\centering
\begin{tabular}{ll}
\hline
\textbf{Parameter} & \textbf{Value} \\
\hline
Batch size & 2 \\
Gradient accumulation steps & 8 \\
Learning rate (initial) & $8 \times 10^{-6}$ \\
LR scheduler & Cosine decay \\
Number of epochs & 1 \\
Image resolution & $1280 \times 1280$ pixels \\
dtype & bfloat16 \\
\hline
\end{tabular}
\caption{Training parameter details for stage 2.}
\label{tab:stage2_hpy}
\end{table}
\subsection*{Training Dynamics}

\begin{figure}[h]
  \centering
  \includegraphics[width=\linewidth]{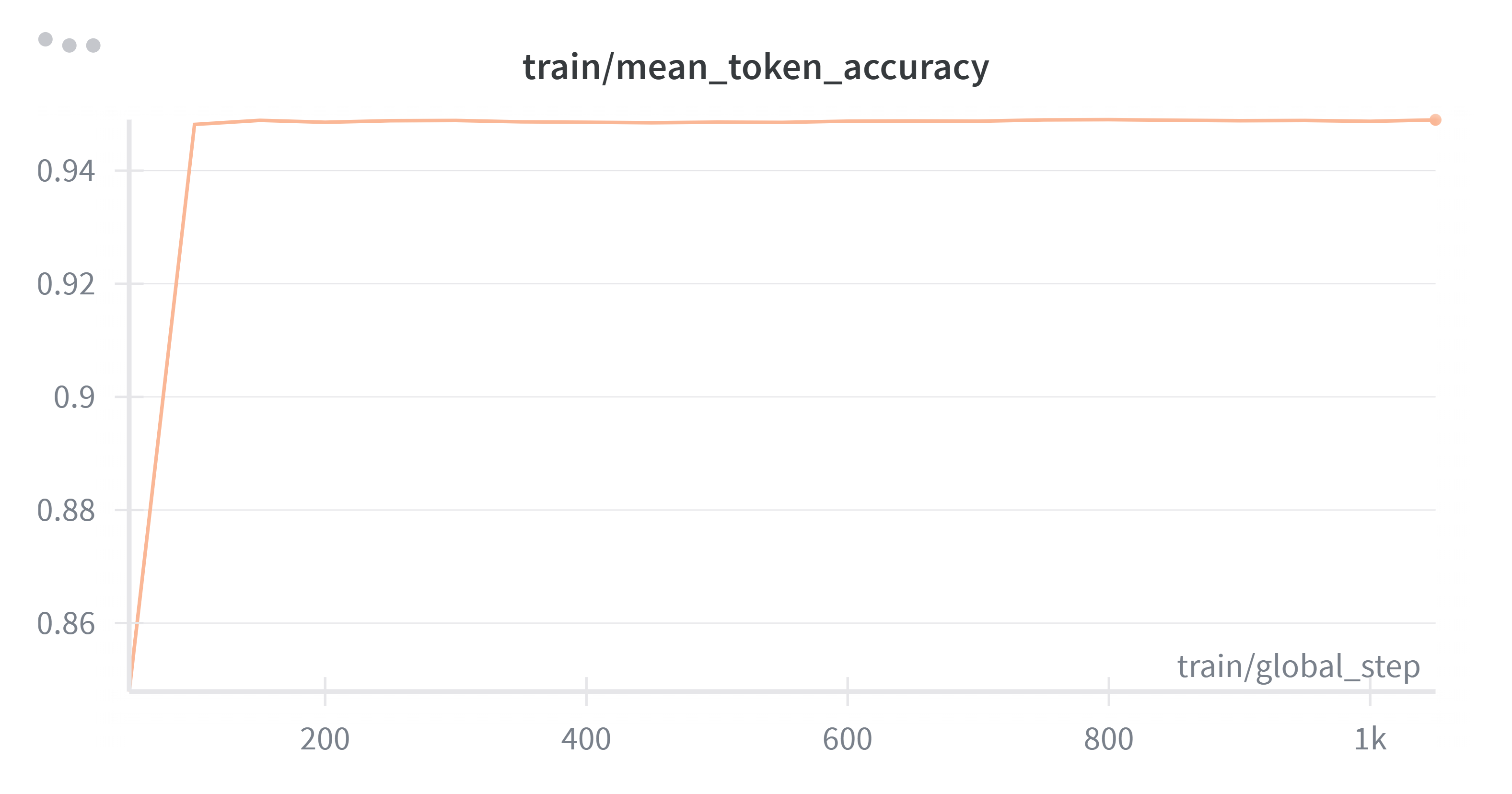}
  \caption{Stage~2 mean token accuracy vs.\ global step (each step = 100 mini-batches). Accuracy improves rapidly from $\sim$0.86 to $\sim$0.95, showing strong convergence and model stability.}
  \label{fig:stage2_acc}
\end{figure}

\begin{figure}[h]
  \centering
  \includegraphics[width=\linewidth]{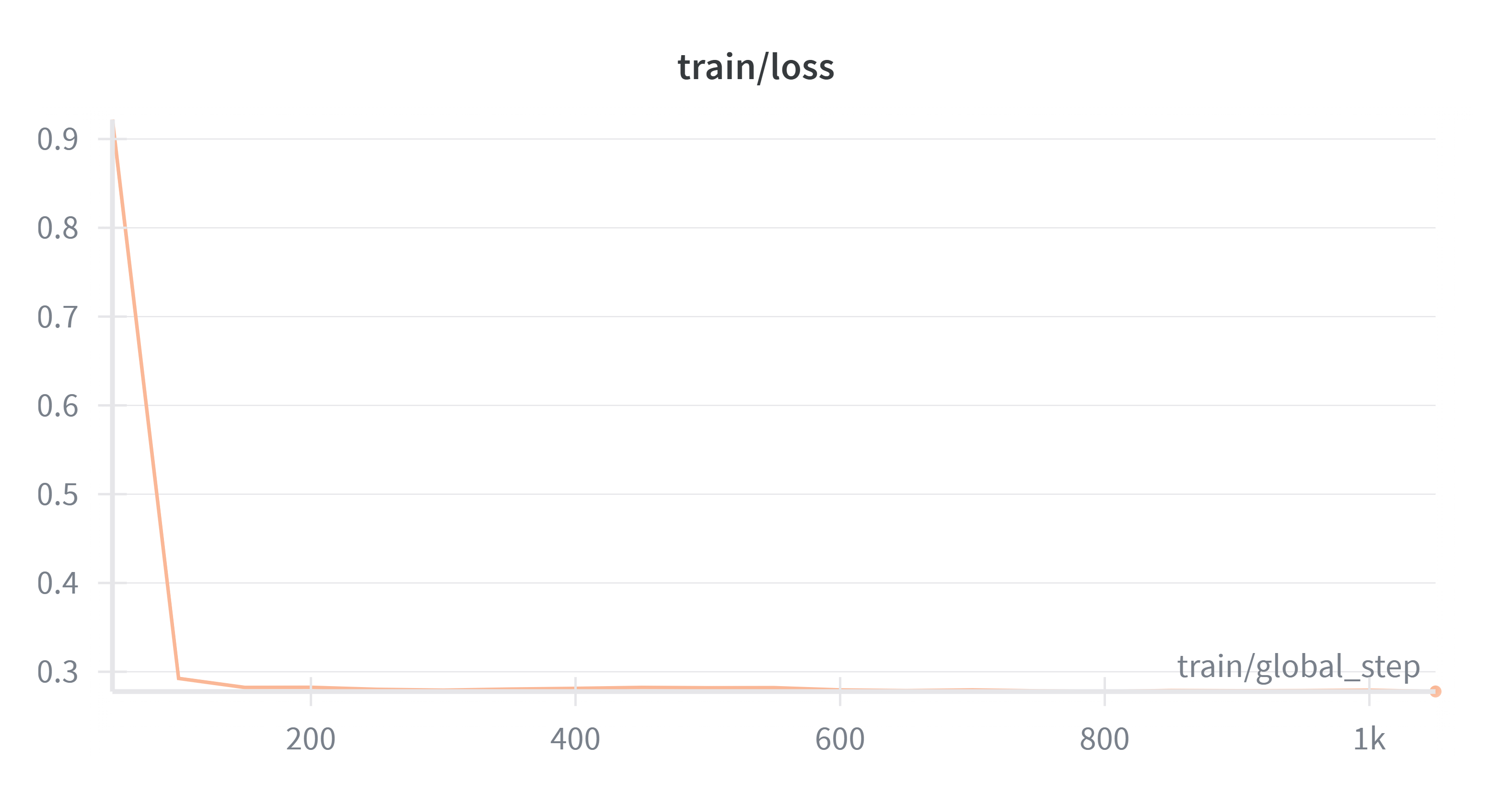}
  \caption{Stage~2 training loss vs.\ global step (each step = 100 mini-batches). Loss decreases from $\sim$0.9 to $\sim$0.3, confirming efficient optimization and stable convergence.}
  \label{fig:stage2_loss}
\end{figure}

\begin{figure}[h]
  \centering
  \includegraphics[width=\linewidth]{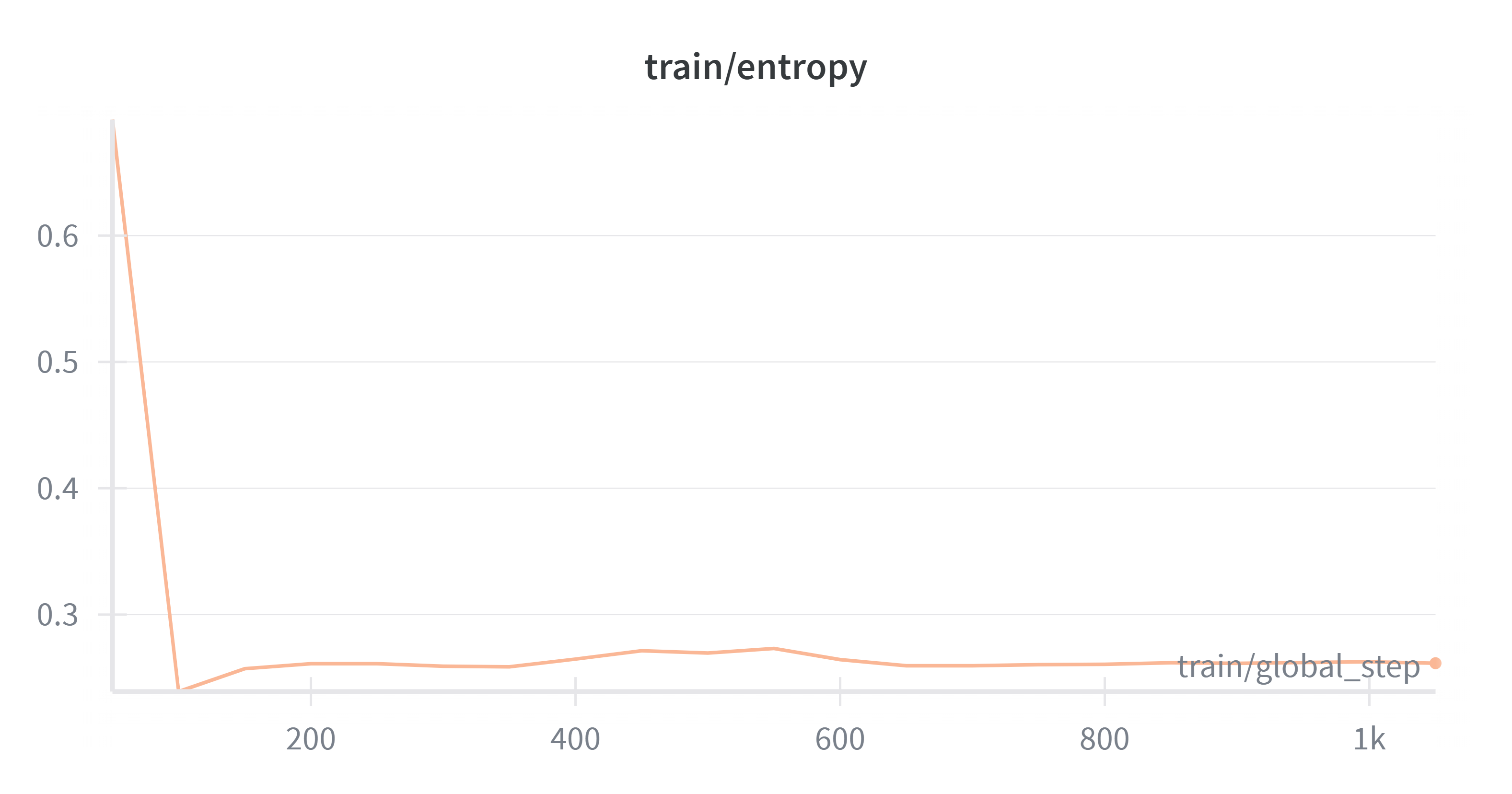}
  \caption{Stage~2 output entropy vs.\ global step (each step = 100 mini-batches). Entropy declines from $\sim$0.65 to $\sim$0.27, reflecting reduced uncertainty and higher confidence in predictions.}
  \label{fig:stage2_entropy}
\end{figure}

During Stage~2, we fine-tuned MedMO on high-resolution ($1280 \times 1280$) medical images using a combination of VQA, grounding, and report-generation datasets. 
Each logged step corresponds to 100 training batches. 
As illustrated in Figures~\ref{fig:stage2_acc}–\ref{fig:stage2_entropy}, the model exhibits rapid convergence and stable learning behavior. 
Mean token accuracy (Fig.~\ref{fig:stage2_acc}) increases sharply from $\sim$0.86 to $\sim$0.95 within the first few hundred steps, indicating strong adaptation to high-resolution visual–textual data. 
Training loss (Fig.~\ref{fig:stage2_loss}) decreases quickly from $\sim$0.9 to $\sim$0.3 and then plateaus, confirming smooth optimization without overfitting. 
Entropy (Fig.~\ref{fig:stage2_entropy}) drops from $\sim$0.65 to $\sim$0.27 and remains steady, showing reduced uncertainty and confident token predictions. 
These results confirm that Stage~2 effectively enhances MedMO’s multimodal alignment and high-resolution spatial reasoning.

\subsection*{Datasets Used}
For Stage~2, we employed datasets emphasizing multimodal reasoning, high-quality medical captions, and spatial grounding. 
The training corpus included a diverse mix of \textbf{VQA-oriented datasets} such as \textit{VQA-Med-2019}, \textit{PubMed-Vision}, \textit{NIH-VQA}, \textit{Quilt-LLaVA-Pretrain}, \textit{MIMIC-Ext-MIMIC-CXR-VQA}, \textit{VQA-RAD}, \textit{PathVQA}, \textit{PMC-VQA}, \textit{SLAKE}, and \textit{CT-RATE}. 
We also incorporated \textbf{report-generation datasets} including \textit{IU-Xray}, \textit{MIMIC-CXR}, \textit{CheXpert}, \textit{CheXpert Plus}, \textit{MEDPIX-ClinQA}, \textit{ROCO}, \textit{ROCO-V2}, and \textit{FairVLMed} to enhance radiology-style narrative generation and image–text consistency. 
Finally, for \textbf{grounding and bounding-box prediction}, we used \textit{NIH Chest X-ray}, \textit{DeepLesion}, \textit{GRAZPEDWRI-DX}, \textit{SLAKE}, \textit{Cell Microscopy (DeepCell, Bacteria, and CTC)}, and \textit{MedSG}, which provide localized annotations for spatial reasoning and fine-grained object detection.

\vspace{0.5em}
This combination allows MedMO to improve fine-grained visual grounding and detailed report synthesis under high-resolution supervision.


\subsection{Stage\,3: Instruction Tuning}

\subsection*{Parameters Details}
We provide detailed experimental settings in Table \ref{tab:stage3_hpy}, which we apply exclusively to training stage 3 MedMO.
\begin{table}[h!]
\centering
\begin{tabular}{ll}
\hline
\textbf{Parameter} & \textbf{Value} \\
\hline
Batch size & 14 \\
Gradient accumulation steps & 2 \\
Learning rate (initial) & $5 \times 10^{-6}$ \\
LR scheduler & Cosine decay \\
Number of epochs & 1 \\
dtype & bfloat16 \\
\hline
\end{tabular}
\caption{Training parameter details for stage 2.}
\label{tab:stage3_hpy}
\end{table}
\subsection*{Training Dynamics}

\begin{figure}[h]
  \centering
  \includegraphics[width=\linewidth]{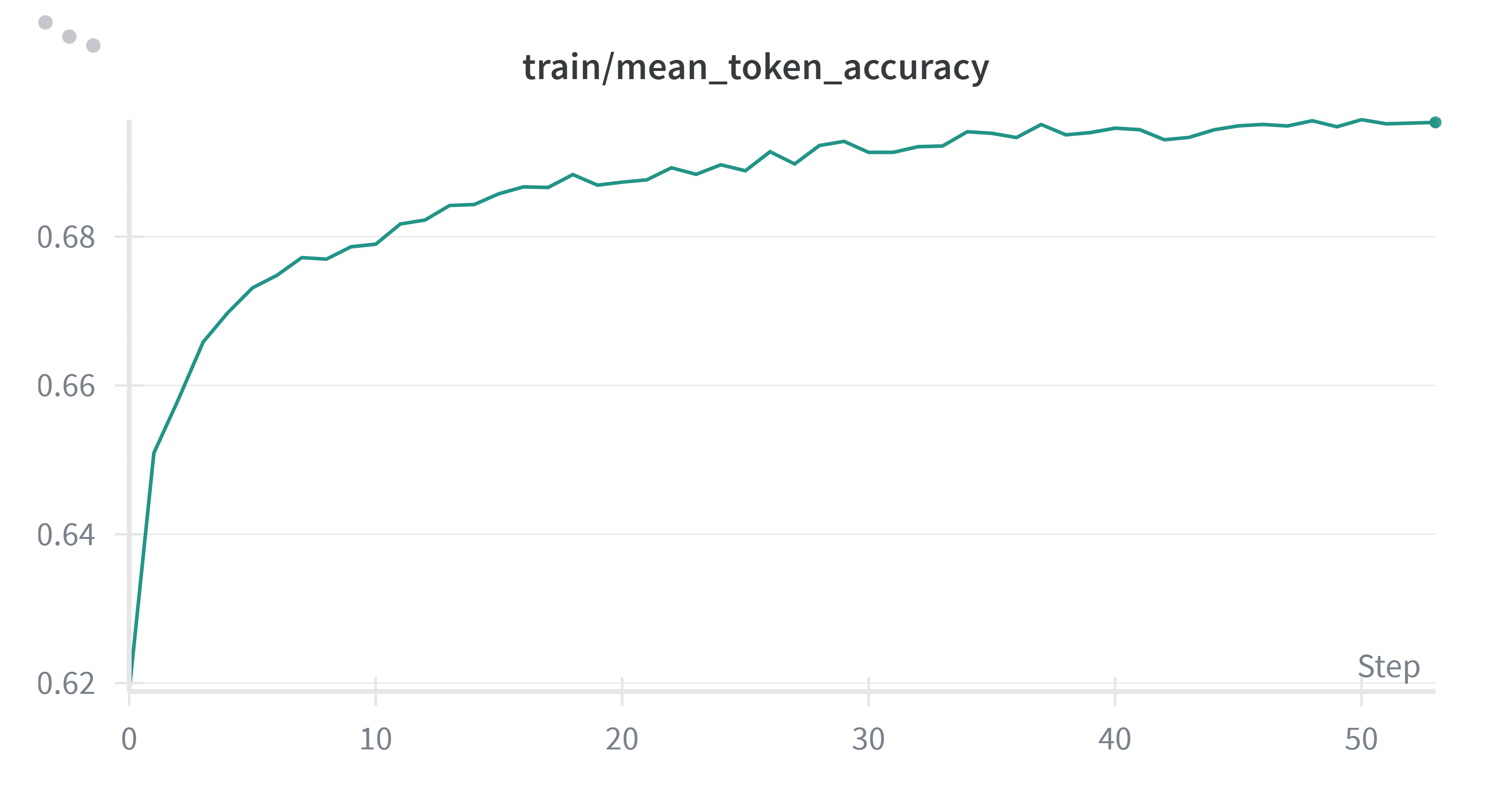}
  \caption{Stage~3 mean token accuracy vs.\ step (each step = 100 mini-batches). Accuracy increases gradually from $\sim$0.62 to $\sim$0.69, indicating improved instruction-following and reasoning.}
  \label{fig:stage3_acc}
\end{figure}

\begin{figure}[h]
  \centering
  \includegraphics[width=\linewidth]{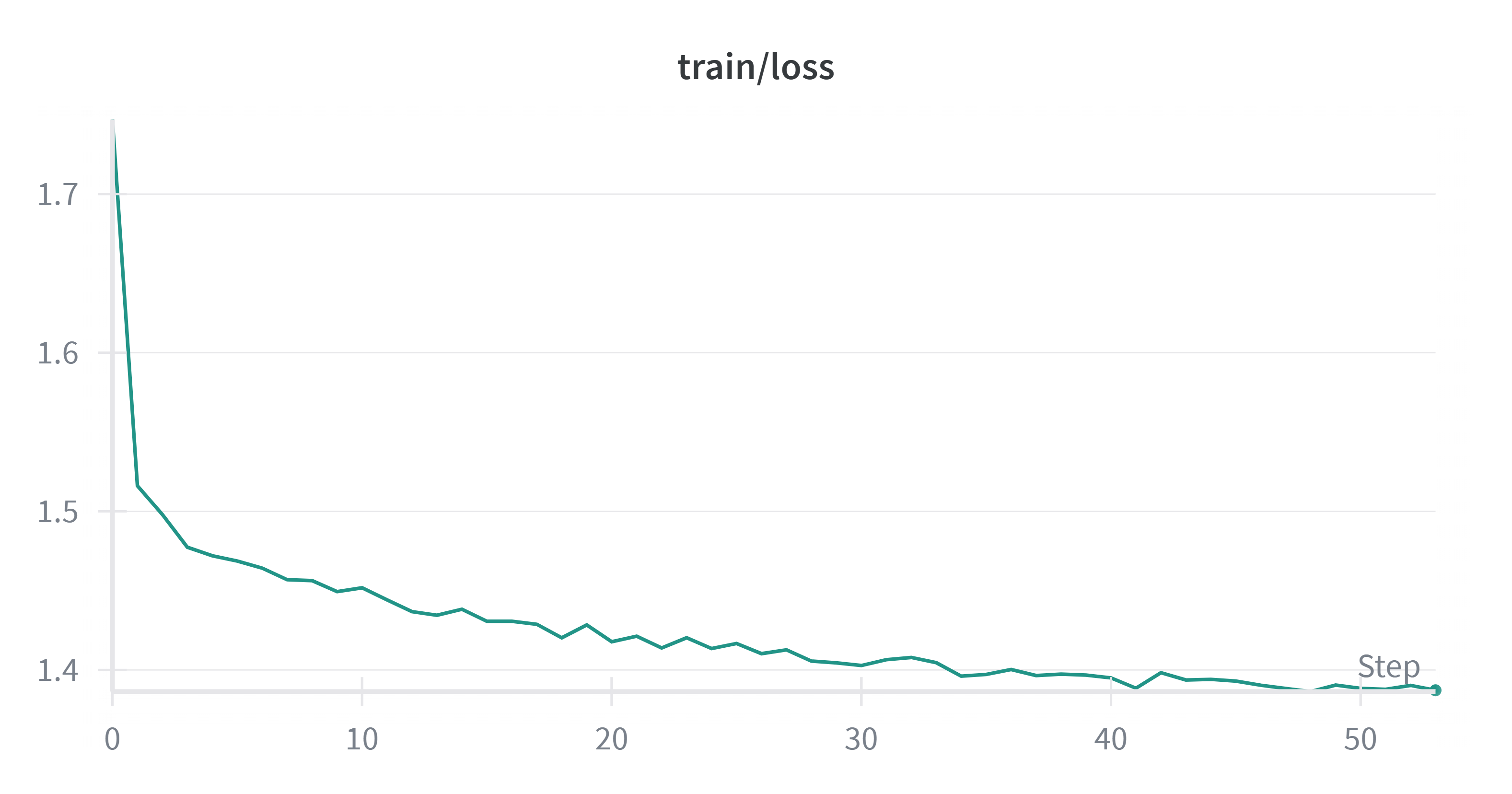}
  \caption{Stage~3 training loss vs.\ step (each step = 100 mini-batches). Loss decreases from $\sim$1.7 to $\sim$1.4, showing smooth convergence and stable optimization.}
  \label{fig:stage3_loss}
\end{figure}

\begin{figure}[h]
  \centering
  \includegraphics[width=\linewidth]{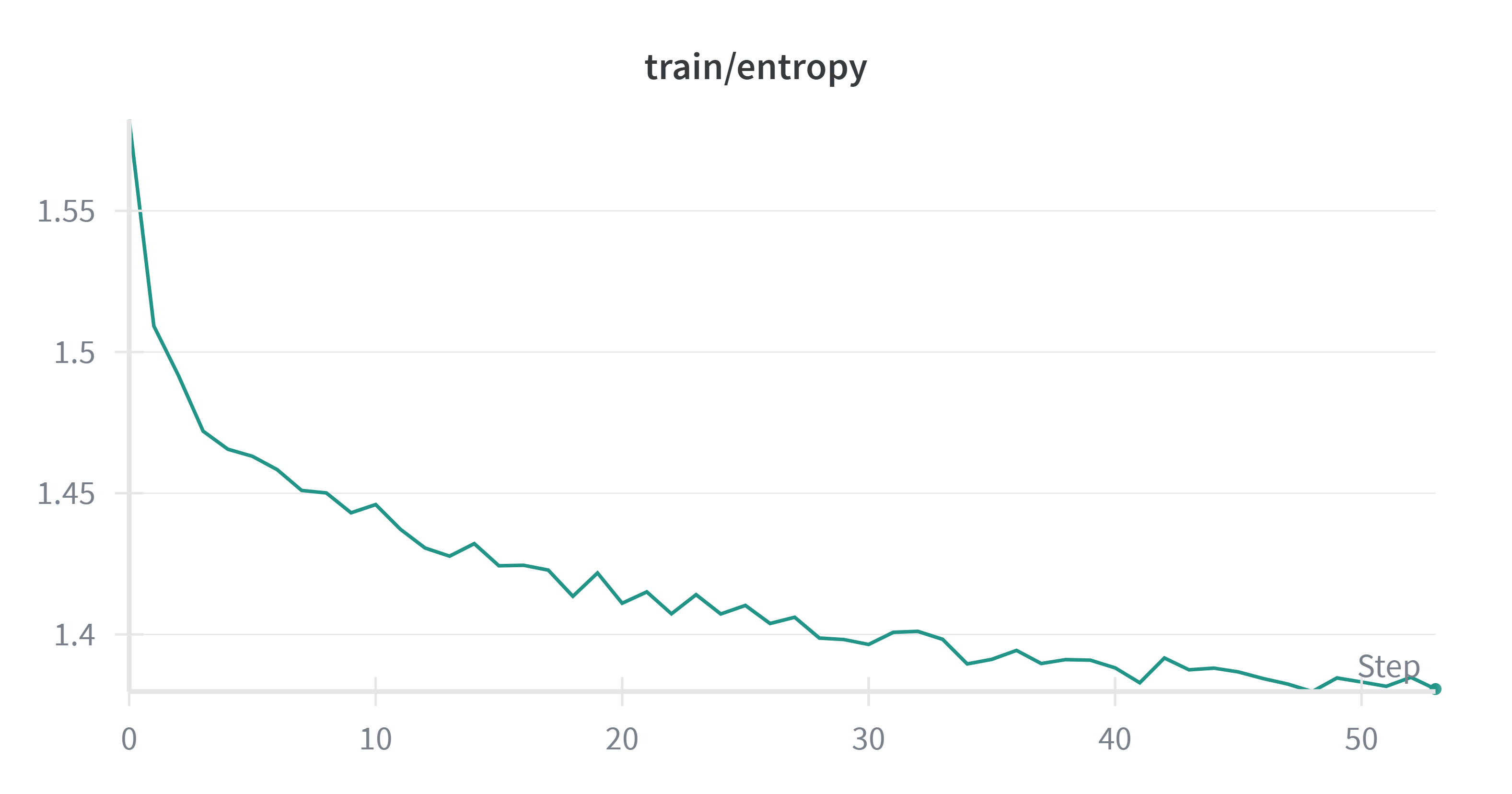}
  \caption{Stage~3 output entropy vs.\ step (each step = 100 mini-batches). Entropy decreases from $\sim$1.55 to $\sim$1.38, reflecting higher model confidence and stable prediction behavior.}
  \label{fig:stage3_entropy}
\end{figure}

Stage~3 focuses on instruction tuning to enhance MedMO’s clinical reasoning, comprehension, and text generation capabilities. 
Each step shown in the plots corresponds to 100 mini-batches. 
As shown in Figures~\ref{fig:stage3_acc}–\ref{fig:stage3_entropy}, the model exhibits smooth and stable convergence. 
Mean token accuracy (Fig.~\ref{fig:stage3_acc}) rises steadily from $\sim$0.62 to $\sim$0.69, demonstrating improved instruction-following and cross-modal reasoning. 
Training loss (Fig.~\ref{fig:stage3_loss}) decreases from $\sim$1.7 to $\sim$1.4 within the first few steps, while entropy (Fig.~\ref{fig:stage3_entropy}) declines from $\sim$1.55 to $\sim$1.38, both indicating effective optimization and improved confidence. 
Overall, Stage~3 consolidates multimodal understanding and instruction-following capabilities with stable convergence and balanced learning dynamics.

\subsection*{Datasets Used}
For Stage~3, we utilized datasets centered on medical instruction-following, comprehension, reasoning, and report summarization. 
The training corpus integrated a broad collection of \textbf{QA and understanding datasets}, including \textit{MedQA}, \textit{PubMedQA}, \textit{PMC-OA}, \textit{MedMCQA}, \textit{PMC-InstructQA}, \textit{MedQuAD}, \textit{Medical-Meadow-MedQA}, \textit{ChatDoctor-HealthCareMagic-100k}, \textit{AlpaCare-MedInstruct-52k}, \textit{ChatDoctor-iCliniq}, \textit{MedReason}, \textit{MIMIC-IV-Ext-BHC}, \textit{Medical-R1-Distill-Data}, \textit{medical-o1-reasoning-SFT}, \textit{Meadow-PubMed-Causal}, \textit{Meadow-Medical-Flashcards}, \textit{Meadow-MediQA}, and \textit{Meadow-Wikidoc}.  
These datasets collectively provide diverse factual, reasoning, and instruction-based supervision across medical, clinical, and biomedical contexts.

In addition, we incorporated \textbf{summarization and clinical reporting datasets} such as \textit{Medical-Meadow-Cord19}, and \textit{mimic-ext-bhc}. 
These datasets focus on long-form radiology and biomedical report synthesis, improving contextual understanding, summarization, and domain-specific narrative generation. 

Together, this combined corpus strengthens MedMO’s instruction-tuned reasoning, factual grounding, and text–image comprehension, enabling robust performance across diverse medical instruction and report-generation scenarios.

\subsection{Stage\,4: Reinforcement Learning (Spatial Grounding)}

\subsection*{Parameters Details}
\begin{itemize}
  \item Reward functions: Label accuracy, bounding‐box IoU ($\Delta$), tag count, and soft‐overlong‐punishment.  
  \item Image resolution: dynamic (no fixed resize or bounding‐box rescaling).  
  \item Epsilon (policy perturbation) = 0.15.  
  \item Epsilon\_high (upper bound) = 0.25.  
  \item Number of training epochs = 2.  
  \item Number of batch size = 2.
  \item Gradient accumulation steps = 4.
  \item Number of generations per prompt = 8.  
  \item Maximum prompt length = 2048 tokens.  
  \item Maximum completion length = 1024 tokens.  
\end{itemize}

\subsection*{Implementation \& Reproducibility Notes}
\begin{itemize}[noitemsep,topsep=2pt,leftmargin=1.5em]
  \item Optimizer: AdamW with default betas (0.9, 0.999) and weight decay = 0.1.  
  \item Warm‐up steps = 10\% of total training steps per stage.   
  \item Seed: All runs initialized with a fixed seed (e.g., 42) per stage; randomness only arises from data shuffling and augmentations.   
\end{itemize}

\subsection*{Training Dynamics}
\begin{figure}[t]
  \centering
  \includegraphics[width=\linewidth]{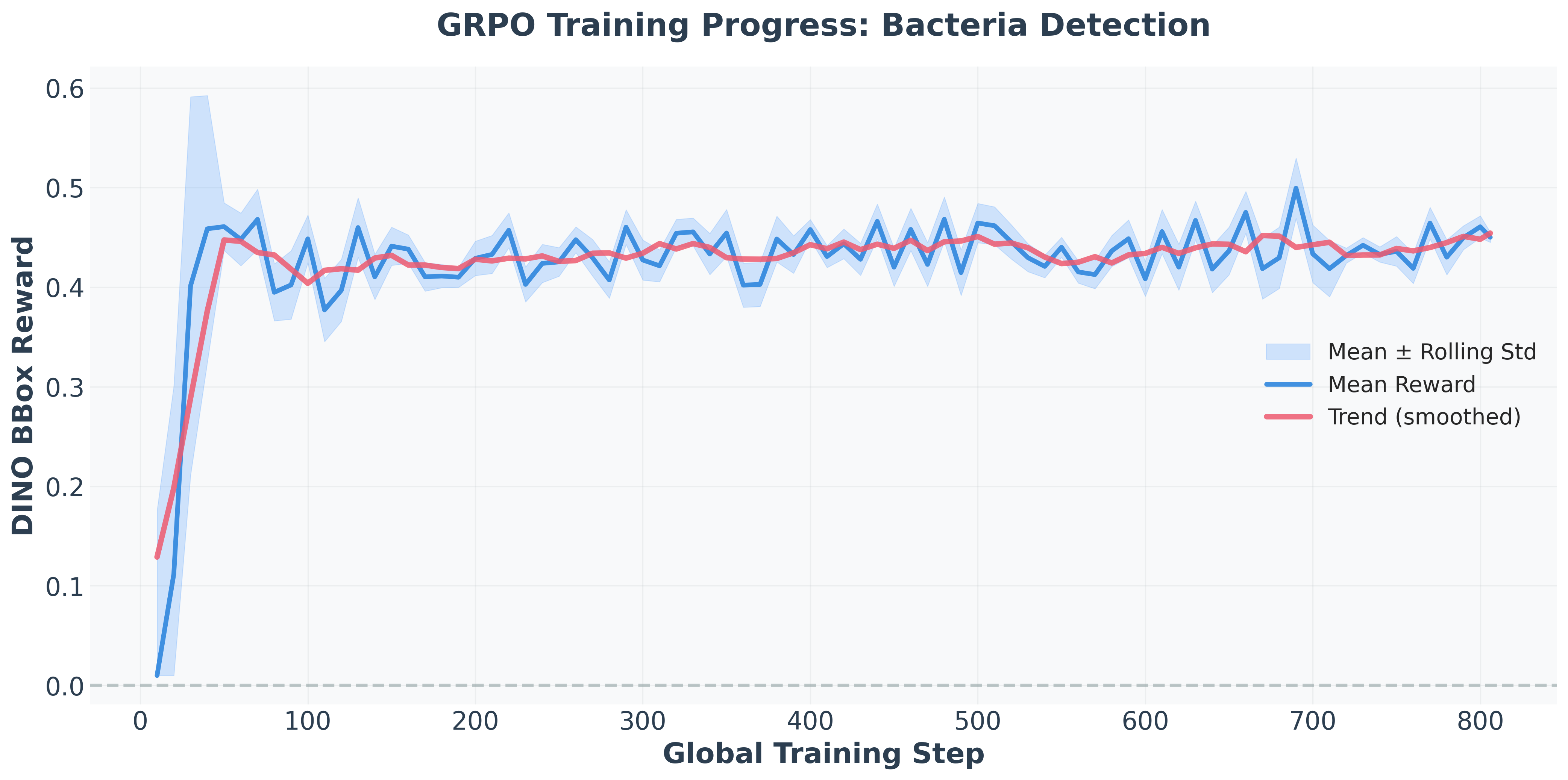}
  \caption{\textbf{DAPO training progress for bounding-box detection.} Mean bounding-box reward (blue) with $\pm$ rolling standard deviation (shaded) and smoothed trend (red). The consistent upward trajectory indicates effective policy optimization and stable improvement in spatial localization accuracy.}
  \label{fig:grpo_training_stats}
\end{figure}

During Stage 4, MedMO was trained with reinforcement learning using the DAPO \cite{yu2025dapo} algorithm to refine its spatial grounding and bounding-box localization capabilities.  
Each global step aggregates multiple rollouts sampled per instruction prompt.  
As shown in Figure \ref{fig:grpo_training_stats}, the bounding-box reward rises sharply from nearly zero to $\sim$0.45 within the first 100 steps, indicating rapid adaptation of the policy to spatial localization signals.  
Beyond this point, the mean reward curve (blue) stabilizes around 0.42–0.45 with moderate oscillations, while the smoothed trend (red) shows a consistent upward trajectory, reflecting incremental performance gains and robust reward optimization.  
The steady variance band (rolling standard deviation) demonstrates that exploration remains controlled throughout training, preventing reward collapse or policy drift.  
Overall, the DAPO stage successfully enhances the model’s spatial precision and stability in bounding-box generation tasks such as bacteria and lesion detection.

\subsection*{Datasets Used}
For Stage~4, we utilized datasets providing explicit spatial supervision and precise bounding-box annotations for medical object detection and grounding tasks.  
These include \textit{NIH Chest X-ray}, \textit{DeepLesion}, \textit{Bacteria Segmentation}, \textit{CTC (Cell Tracking Challenge)}, \textit{SLAKE}, \textit{GRAZPEDWRI-DX}, and \textit{MedSG}, which collectively cover anatomical structures, lesions, and microscopic cellular regions.  
The DAPO objective leverages bounding-box IoU and label-accuracy rewards derived from these datasets to iteratively refine spatial alignment and improve localization precision.  
This stage significantly enhances MedMO’s visual grounding ability, leading to robust disease localization and fine-grained spatial reasoning across diverse medical modalities.

\section{Dataset Collection}

We curated a unified multimodal corpus comprising \textbf{45 datasets} spanning radiology, pathology, ophthalmology, dermatology, and surgical imaging, totaling more than \textbf{26M samples}. 
At the core lies the \textbf{MedTrinity} dataset~ \cite{xie2025medtrinity}, which contributes \textbf{18.5M} publicly available instruction-following pairs. 
This large-scale collection integrates both image–text and text-only medical data, enabling tasks such as captioning, visual question answering (VQA), clinical reasoning, and visual grounding.

The model was trained through four progressive stages. 
In \textbf{Stage~1}, we used the MedTrinity dataset to establish foundational multimodal understanding across diverse imaging modalities. 
\textbf{Stage~2} incorporated additional VQA, grounding, and captioning datasets, and trained the model with high-resolution medical images to enhance visual reasoning and fine-grained spatial grounding.

\textbf{Stage~3} focused on medical text-only instruction data to strengthen clinical knowledge and language understanding. 
Finally, \textbf{Stage~4} employed reinforcement learning with bounding-box supervision to further refine localization and grounding capabilities.

The datasets encompass a broad spectrum of imaging modalities (X-ray, CT, MRI, ultrasound, optical, and nuclear imaging) and biological systems (chest, brain, heart, liver, kidney, eye, colon, and tissue), ensuring comprehensive anatomical and modality coverage. 
For grounding supervision, we incorporated datasets containing bounding-box annotations, including \emph{NIH Chest X-ray}, \emph{DeepLesion}, \emph{Bacteria}, \emph{Wrist X-ray (boneanomaly, fracture etc.)}, \emph{CT}, and \emph{Cell Microscopy (DeepCell)}. 
This diverse corpus collectively supports robust multimodal alignment, spatial reasoning, and medical instruction tuning.

Table~\ref{tab:dataset_collection} summarizes the datasets used in MedMO’s training pipeline, grouped according to their primary role in each stage.

\textcolor{blue}{\textit{Note.} Several other publicly available datasets such as \emph{TCGA}  \cite{tomczak2015review}, \emph{VALSET}  \cite{tolkach2023artificial}, \emph{MAMA-MIA}  \cite{garrucho2024mama}, \emph{LLD-MMRI}  \cite{LLD-MMRI}, \emph{CPD} \cite{wagner2023semantic}, \emph{CISC} \cite{gamper2020pannuke}, \emph{CT-RATE} \cite{hamamci2024generatect}, \emph{KIPA22} \cite{yongchengyao_kipa22_2025}, and \emph{PTCGA}  \cite{kawai2023large} are already included in MedTrinity and were not trained on separately.}

\begin{table*}[ht]
\centering
\caption{
Overview of datasets used in \textbf{MedMO} training. 
Datasets are grouped by category, each contributing to distinct training objectives such as image captioning, multimodal and text-based instruction tuning, and spatial grounding.
}
\label{tab:dataset_collection}
\renewcommand{\arraystretch}{1.2}
\resizebox{\textwidth}{!}{
\begin{tabular}{
    >{\raggedright\arraybackslash}p{3.5cm} 
    >{\raggedright\arraybackslash}p{7cm} 
    >{\raggedright\arraybackslash}p{5cm}
}
\toprule
\textbf{Category} & \textbf{Datasets} & \textbf{Purpose / Usage} \\
\midrule
\textbf{Medical Caption Data} &
\textit{MedTrinity}~\cite{xie2025medtrinity}, \textit{IU-Xray}~\cite{chen-emnlp-2020-r2gen}, \textit{MIMIC-CXR}~\cite{johnson2019mimic}, \textit{CheXpert}~\cite{irvin2019chexpert}, \textit{CheXpert Plus}~\cite{chambon2024chexpert}, \textit{MEDPIX-ClinQA}~\cite{siragusa24072}, \textit{ROCO}~\cite{Pelka2018ROCO}, \textit{ROCO-V2}~\cite{ruckert2024rocov2}, \textit{FairVLMed}~\cite{luo2024fairclip} &
Used for large-scale image–text alignment, caption-based supervision, and radiology-style report modeling across diverse imaging modalities. \\
\midrule
\textbf{Medical Multimodal Instruction Data} &
\textit{VQA-Med-2019}~\cite{abacha2019vqa}, \textit{PubMed-Vision}~\cite{chen2024huatuogpt}, \textit{NIH-VQA}~\cite{sarrouti2020nlm}, \textit{Quilt-LLaVA-Pretrain}~\cite{seyfioglu2024quilt}, \textit{MIMIC-Ext-MIMIC-CXR-VQA}~\cite{bae2024mimic}, \textit{VQA-RAD}~\cite{lau2018dataset}, \textit{PathVQA}~\cite{he2020pathvqa}, \textit{PMC-VQA}~\cite{zhang2023pmc}, \textit{SLAKE}~\cite{liu2021slake}, \textit{CT-RATE}~\cite{brodoefel2008dual} &
Facilitates multimodal instruction tuning for VQA, diagnosis, reasoning, and clinical summarization, improving image–text comprehension and task-driven responses. \\
\midrule
\textbf{Medical Text Instruction Data} &
\textit{MedQA}~\cite{yang2024llm}, \textit{PubMedQA}~\cite{jin2019pubmedqa}, \textit{PMC-OA}~\cite{lin2023pmc}, \textit{MedMCQA}~\cite{pal2022medmcqa}, \textit{PMC-InstructQA}~\cite{zhang2023pmc}, \textit{MedQuAD}~\cite{BenAbacha-BMC-2019}, \textit{Medical-Meadow-MedQA}~\cite{jin2020disease}, \textit{ChatDoctor-HealthCareMagic-100k}~\cite{lavita_chatdoctor_healthcaremagic_100k_2023}, \textit{AlpaCare-MedInstruct-52k}~\cite{zhang2023alpacareinstructiontuned}, \textit{ChatDoctor-iCliniq}~\cite{lhoest-etal-2021-datasets}, \textit{MedReason}~\cite{wu2025medreason}, \textit{MIMIC-IV-Ext-BHC}~\cite{aali2024mimic}, \textit{Medical-R1-Distill-Data}~\cite{chen2024huatuogpto1medicalcomplexreasoning}, \textit{medical-o1-reasoning-SFT}~\cite{chen2024huatuogpto1medicalcomplexreasoning}, \textit{Meadow-PubMed-Causal}~\cite{yu-etal-2019-detecting}, \textit{Meadow-Medical-Flashcards}~\cite{medalpaca_medical_meadow_medical_flashcards_2023}, \textit{Meadow-MediQA}~\cite{ratliff1985meadows}, \textit{Meadow-Wikidoc}~\cite{medalpaca_medical_meadow_wikidoc_2023}, \textit{Medical-Meadow-Cord19}~\cite{wang-etal-2020-cord}, \textit{mimic-ext-bhc}~\cite{aali2025mimic} &
Provides text-only instruction and QA supervision to enhance factual reasoning, language understanding, and medical knowledge grounding across clinical and biomedical contexts. \\
\midrule
\textbf{Medical Bounding Box Data} &
\textit{NIH Chest X-ray}~\cite{filice2020crowdsourcing}, \textit{DeepLesion}~\cite{yan2018deeplesion}, \textit{GRAZPEDWRI-DX}~\cite{nagy2022pediatric}, \textit{SLAKE}~\cite{liu2021slake}, \textit{Cell Microscopy (DeepCell, Bacteria, CTC)}~\cite{van2018spatially}, \textit{MedSG}~\cite{yue2025medsg} &
Provides explicit spatial grounding and disease-localization supervision with bounding-box annotations, enabling fine-grained object detection and improved spatial reasoning in medical imagery. \\
\bottomrule
\end{tabular}
}
\end{table*}


\section{Qualitative Results}
\begin{figure*}[h!]
  \centering
  \fcolorbox{black!20}{white}{%
    \includegraphics[width=0.98\textwidth]{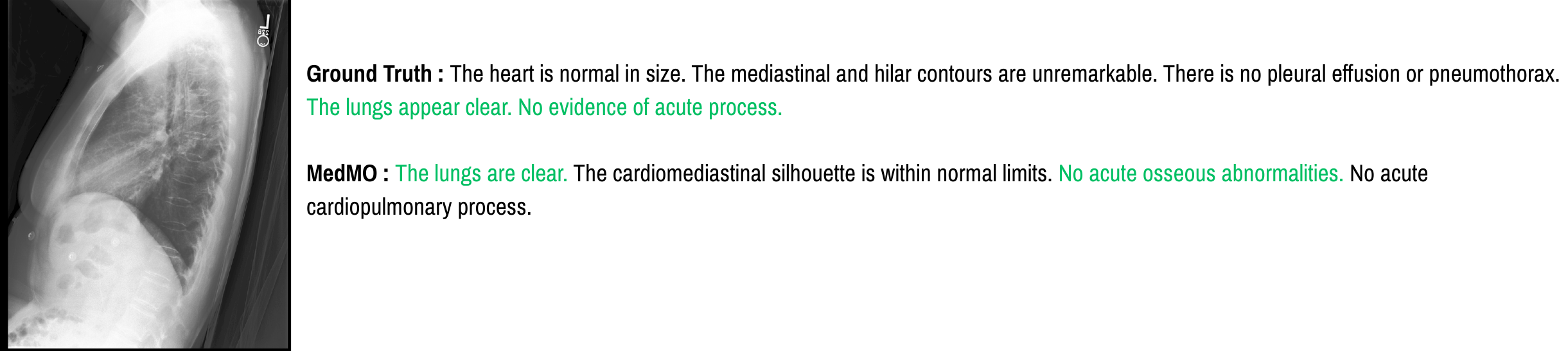}
  }
  \caption{\textbf{Qualitative result example \#1.} We show model predictions compared against ground truth annotations. The input medical image is displayed on the left, with corresponding text outputs on the right. Correct predictions are highlighted (highlighted in \textcolor{green}{green}) to demonstrate alignment with clinical ground truth, while differences indicate areas for potential improvement.}
  \label{fig:supply1}
\end{figure*}

\begin{figure*}[h!]
  \centering
  \fcolorbox{black!20}{white}{%
    \includegraphics[width=0.98\textwidth]{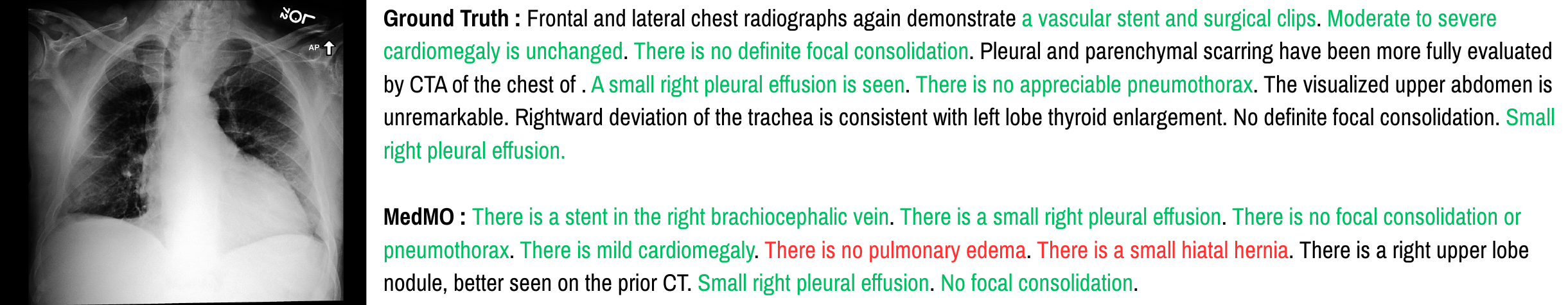}
  }
  \caption{\textbf{Qualitative result example \#2.} We show model predictions compared against ground truth annotations. The input medical image is displayed on the left, with corresponding text outputs on the right. Correct predictions are highlighted (highlighted in \textcolor{green}{green}) to demonstrate alignment with clinical ground truth, while differences indicate areas for potential improvement.}
  \label{fig:supply2}
\end{figure*}

\begin{figure*}[h!]
  \centering
  \fcolorbox{black!20}{white}{%
    \includegraphics[width=0.98\textwidth]{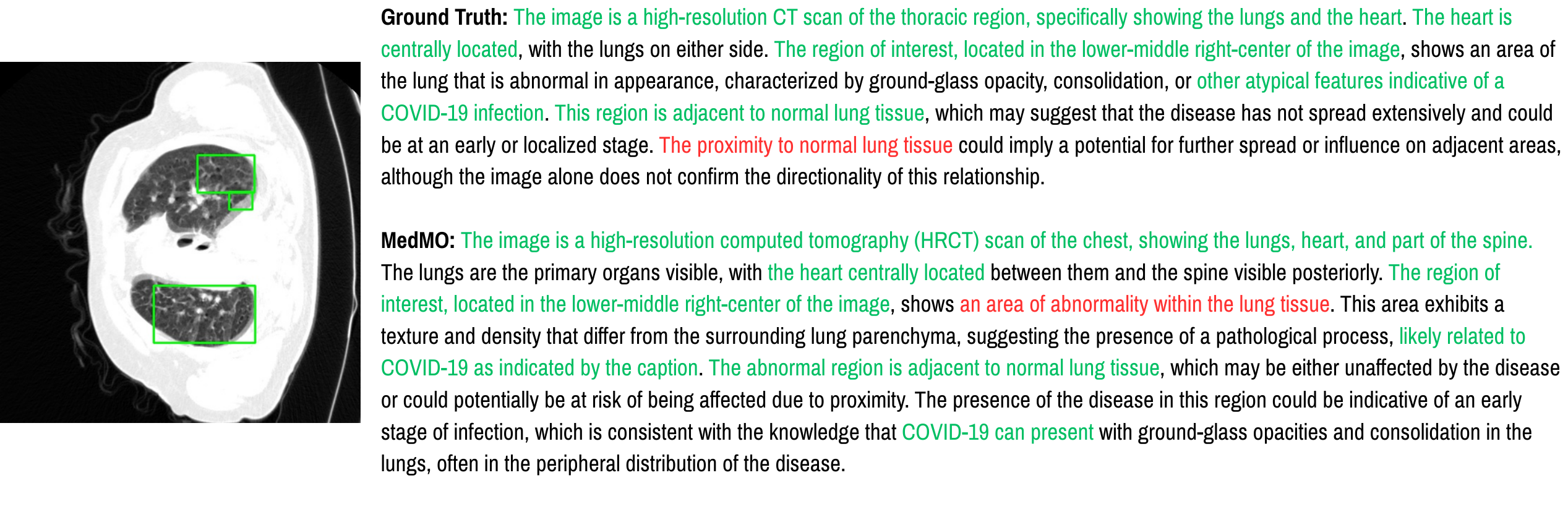}
  }
  \caption{\textbf{Qualitative result example \#3.} We show model predictions compared against ground truth annotations. The input medical image is displayed on the left, with corresponding text outputs on the right. Correct predictions are highlighted (highlighted in \textcolor{green}{green}) to demonstrate alignment with clinical ground truth, while differences indicate areas for potential improvement.}
  \label{fig:supply3}
\end{figure*}

\begin{figure*}[h!]
  \centering
  \fcolorbox{black!20}{white}{%
    \includegraphics[width=0.98\textwidth]{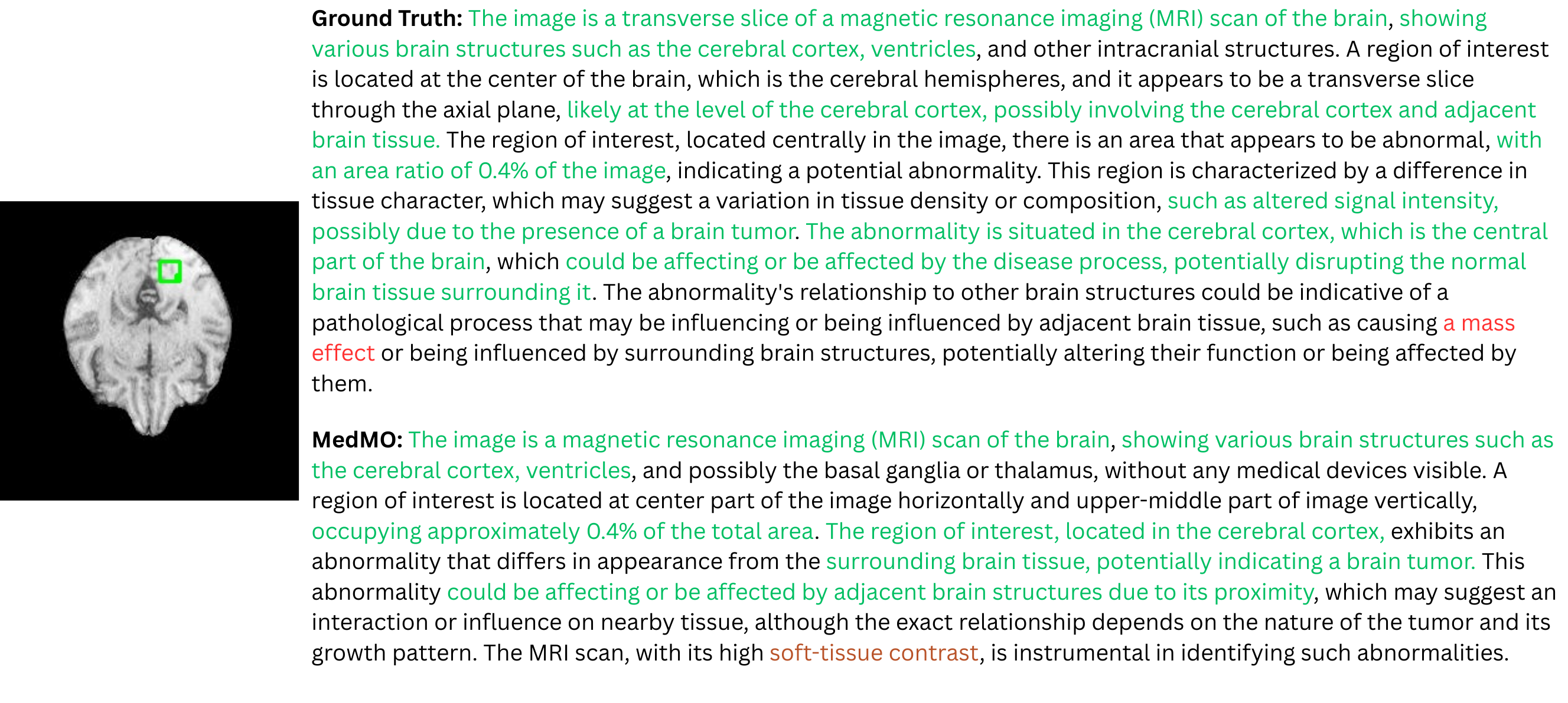}
  }
  \caption{\textbf{Qualitative result example \#4.} We show model predictions compared against ground truth annotations. The input medical image is displayed on the left, with corresponding text outputs on the right. Correct predictions are highlighted (highlighted in \textcolor{green}{green}) to demonstrate alignment with clinical ground truth, while differences indicate areas for potential improvement.}
  \label{fig:supply4}
\end{figure*}

To complement the quantitative analyses presented in the main text, Figures~\ref{fig:supply1}–\ref{fig:supply4} provide qualitative insights into our method's performance across diverse medical imaging scenarios. These visualizations illustrate representative predictions, highlighting both successful cases and challenging examples under varied clinical conditions.

\section{Overall Training Summary}
Across the four stages, MedMO progressively improves from general multimodal alignment (Stage 1) 
to high-resolution spatial reasoning and grounding (Stage 2), instruction-tuned language understanding (Stage 3), 
and reinforcement-driven grounding refinement (Stage 4). 
Together, these stages establish a robust, domain-aware foundation model for diverse medical imaging tasks.

